\documentclass[11pt, letterpaper]{arxiv}
\usepackage[all]{hypcap}
\usepackage[comma,numbers,sort,compress]{natbib}
\usepackage{hyperref}[citecolor=magenta]
\usepackage[capitalize,noabbrev]{cleveref}
\hypersetup{
    colorlinks = true,
    citecolor = {magenta},
}

\usepackage{microtype}
\usepackage{graphicx}
\usepackage{subfigure}
\usepackage{booktabs} 

\usepackage{xcolor}
\usepackage{colortbl}
\usepackage{algorithm}
\usepackage{algpseudocode}
\usepackage{setspace}

\def\Snospace~{\S{}}

\definecolor{coco1}{HTML}{D9E4EC}
\definecolor{coco2}{HTML}{B7CFDC}
\definecolor{coco3}{HTML}{6AABD2}
\definecolor{coco4}{HTML}{385E72}
\hypersetup{
    colorlinks=true,
    linkcolor=coco3,
    filecolor=coco4,      
    urlcolor=coco3,
    citecolor=coco3,
}

\newcommand{\ourmethod}{\texttt{GIANTS-4B}}

\newcommand{\ourbenchmark}{\textsc{GiantsBench}}
\newcommand{\gemtwoflash}{\texttt{gemini-2.5-flash}}
\newcommand{\gemtwopro}{\texttt{gemini-2.5-pro}}
\newcommand{\gemthreepro}{\texttt{gemini-3-pro}}
\newcommand{\gemthreeflash}{\texttt{gemini-3-flash}}
\newcommand{\qwenfourb}{\texttt{Qwen3-4B}}
\newcommand{\qwenfourteenb}{\texttt{Qwen3-14B}}

\usepackage{float}
\usepackage{caption}

\usepackage{amsmath}
\usepackage{amssymb}
\usepackage{mathtools}
\usepackage{amsthm}

\usepackage{placeins}

\usepackage{fleqn, tabularx}

\usepackage{float}

\usepackage[most,skins,theorems]{tcolorbox}
\tcbset{
  aibox/.style={
    width=\linewidth,
    top=7pt,
    bottom=2pt,
    colback=blue!6!white,
    colframe=black,
    colbacktitle=black,
    enhanced,
    center,
    attach boxed title to top left={yshift=-0.1in,xshift=0.15in},
    boxed title style={boxrule=0pt,colframe=white,},
  }
}
\newtcolorbox{AIbox}[2][]{aibox,title=#2,#1}

\usepackage{wrapfig}
\definecolor{lightblue}{rgb}{0.22,0.45,0.70}
\usepackage{fleqn, tabularx}

\definecolor{rliableolive}{HTML}{BBCC33}
\definecolor{rliableblue}{HTML}{77AADD}
\definecolor{rliablered}{HTML}{EE8866}

\usepackage{crossreftools}
\usepackage[suppress]{color-edits}

\usepackage{dsfont}
\usepackage{nicefrac}
\newtcolorbox{analysisbox}[1][]{
    enhanced jigsaw,
    colback=white,
    colframe=blue!75!black,
    fonttitle=\bfseries,
    boxsep=5pt,
    left=5pt,
    right=5pt,
    top=5pt,
    bottom=5pt,
    title=#1,
}
\definecolor{editInitialResponse}{RGB}{255, 235, 156} 
\definecolor{editBacktrack}{RGB}{0, 0, 139}
\definecolor{editRevisedResponse}{RGB}{255, 182, 193}

\usepackage{pifont}
\usepackage{soul}
\definecolor{highlightmistake}{RGB}{255, 179, 179} 
\definecolor{highlightcorrect}{RGB}{179, 255, 179}

\usepackage[capitalize,noabbrev]{cleveref}

\theoremstyle{plain}

\theoremstyle{definition}

\theoremstyle{remark}

\usepackage[textsize=tiny]{todonotes}
\usepackage{multirow}
\usepackage{enumitem}

\usepackage{amsmath,amsfonts,bm}

\def\eqref#1{Eq.~\ref{#1}}

\def\1{\bm{1}}

\DeclareMathAlphabet{\mathsfit}{\encodingdefault}{\sfdefault}{m}{sl}
\SetMathAlphabet{\mathsfit}{bold}{\encodingdefault}{\sfdefault}{bx}{n}

\newtcolorbox{promptbox}[2][]{  
listing only,
enhanced,
breakable,
colback=blue!6!white,
colframe=black,
fontupper=\ttfamily,
title=#2,
#1}

\newcommand{\methodname}{{\texttt{GIANTS}}}

\title{\methodname{}: Generative Insight Anticipation from Scientific Literature}

\author[*1]{Joy He-Yueya}
\author[*1]{Anikait Singh}
\author[1]{Ge Gao}
\author[1]{Michael Y. Li}
\author[2]{Sherry Yang}
\author[1]{Chelsea Finn}
\author[1]{Emma Brunskill}
\author[1]{Noah D. Goodman}

\affil[1]{Stanford University}
\affil[2]{New York University}
\affil[*]{Equal contribution}

\correspondingauthor{heyueya@cs.stanford.edu, anikait@stanford.edu. \newline \textbf{Project Website:} \textnormal{\url{https://giants-insights.github.io/}}}

\begin{document}

\maketitle

Scientific breakthroughs often emerge from synthesizing prior ideas into novel contributions. While language models (LMs) show promise in scientific discovery, their ability to perform this targeted, literature-grounded synthesis remains underexplored. We introduce \emph{insight anticipation}, a generation task in which a model predicts a downstream paper's core insight from its foundational parent papers. To evaluate this capability, we develop \ourbenchmark{}, a benchmark of $17k$ examples across eight scientific domains, where each example consists of a set of parent papers paired with the core insight of a downstream paper. We evaluate models using an LM judge that scores similarity between generated and ground-truth insights, and show that these similarity scores correlate with expert human ratings. Finally, we present \ourmethod{}, an LM trained via reinforcement learning (RL) to optimize insight anticipation using these similarity scores as a proxy reward. Despite its smaller open-source architecture, \ourmethod{} outperforms proprietary baselines and generalizes to unseen domains, achieving a 34\% relative improvement in similarity score over \gemthreepro{}. Human evaluations further show that \ourmethod{} produces insights that are more conceptually clear than those of the base model. In addition, SciJudge-30B, a third-party model trained to compare research abstracts by likely citation impact, predicts that insights generated by \ourmethod{} are more likely to lead to higher citations, preferring them over the base model in 68\% of pairwise comparisons. We release our code, benchmark, and model to support future research in automated scientific discovery.

\begin{flushright}
``If I have seen further [than others], it is by standing on the shoulders of giants.'' \\--- Isaac Newton
\end{flushright}

\section{Introduction}
Language Models (LMs) are becoming useful tools for scientific discovery~\citep{karpathy2026autoresearch}. Recent work has shown promising results, from virtual LM teams designing SARS-CoV-2 nanobody binders~\citep{swanson2025virtual} to models proposing NLP research directions that human experts judge to be novel~\citep{si2024can}. However, many of these successes rely heavily on prompting frontier models pre-trained on massive text corpora. In contrast, human researchers often achieve breakthroughs through a far more data-efficient process: synthesizing profound insights from a small set of prior works. Existing LMs still struggle to reliably generate hypotheses or \emph{insights} of true impact and value~\citep{si2025ideation}, often due to the lack of diversity and feasibility~\citep{wadhwa2026createtestingllmsassociative}. 

To bridge this gap, we propose the task of \emph{insight anticipation}: given a small set of prior papers, can a model reconstruct the core insight of a downstream paper that builds on them? Unlike open-ended research ideation, this setting evaluates targeted synthesis grounded in specific scientific literature. This framing is motivated by the classical view of scientific progress as `\emph{standing on the shoulders of giants,}' where new contributions emerge by building upon prior foundations. Under this view, the challenge of automated discovery decomposes into two subproblems: (1) \textit{parent selection}, which involves identifying the relevant prior works to build upon, and (2) \textit{insight generation}, which involves synthesizing those works into a novel hypothesis.

In this work, we focus exclusively on the second problem. We assume that the parent papers are provided by an oracle literature selection criterion and ask a more targeted feasibility question: if the relevant lineage of prior work is known, can a model effectively predict the next conceptual leap? This controlled setup isolates the synthesis of research insights from the retrieval of prior work and allows us to directly test whether meaningful literature-grounded insight generation is possible at all. Framed this way, \emph{insight anticipation} is not only a prediction task but also a potential training signal for scientific reasoning: learning to anticipate the downstream insights implied by a lineage of parent works may help models internalize patterns of progression, combination, and abstraction that underlie strong human insight synthesis.

\begin{figure}[t]
    \centering
    \includegraphics[width=\linewidth]{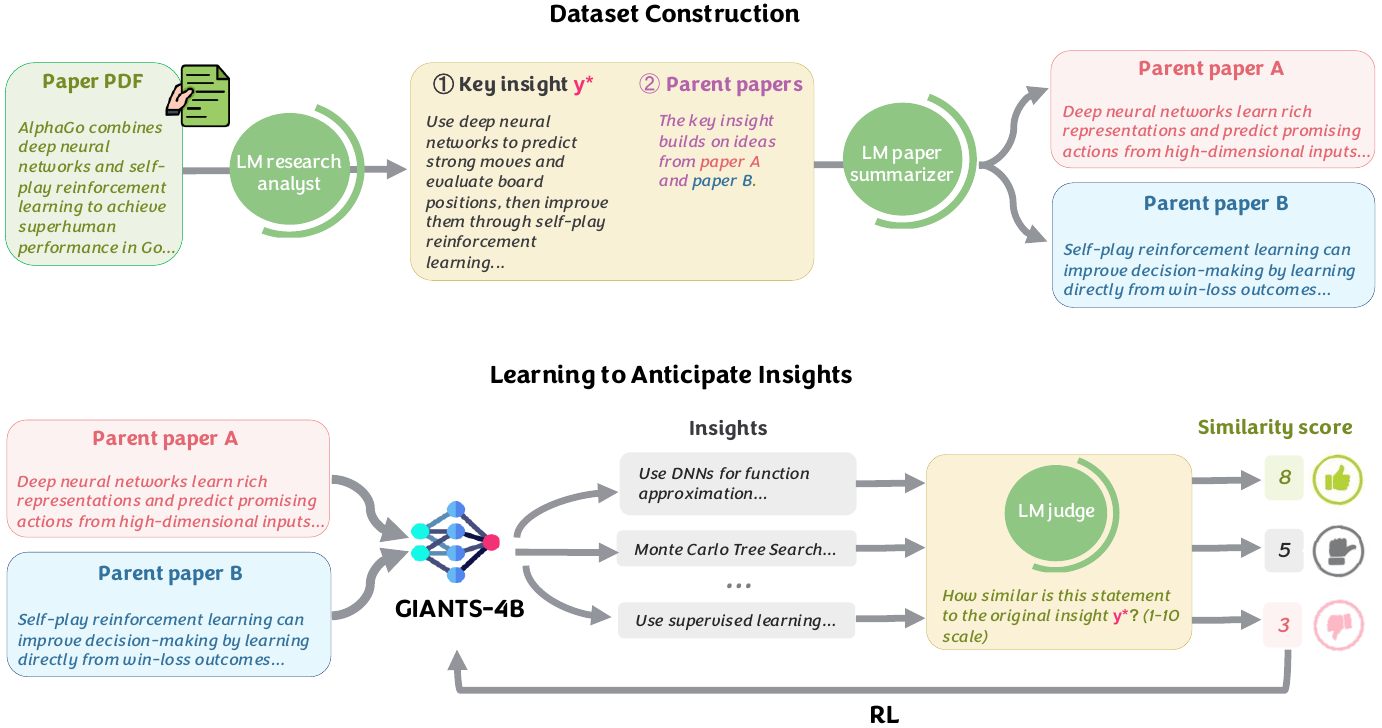}
    \vspace{-3mm}
    \caption{\footnotesize \textbf{Overview of \ourbenchmark{} and \ourmethod{}.} \textit{(top)} For dataset construction, we use an LM as a research analyst to process each paper PDF to identify two parent papers whose ideas are synergistically combined to produce the paper's key insight, which is extracted as the ground-truth target $y^*$. A second LM summarizes each parent paper. \textit{(bottom)} Given the two parent summaries, the model generates candidate insights, which an LM judge scores by similarity to $y^*$ on a 1–10 scale. These scores serve as a proxy reward for RL training, teaching the model to anticipate insights that more closely match real downstream papers.}
    \label{fig:giants_fig1}
    \vspace{-3mm}
\end{figure}

To evaluate insight anticipation models, we develop \ourbenchmark{}, a large-scale benchmark for testing the ability of models to synthesize two prior papers and derive the core insight of a downstream paper that builds on them (Figure~\ref{fig:giants_fig1}, top). In contrast to a recent benchmark that evaluates whether models can answer literature synthesis questions by identifying relevant papers and generating long-form responses with citations \citep{asai2026synthesizing}, \ourbenchmark{} assesses whether a model can combine two parent papers synergistically to derive insights that lead to a subsequent paper. \ourbenchmark{} contains $17k$ examples drawn from Computer Science, Economics, Electrical Engineering, Mathematics, Physics, Quantitative Biology, Quantitative Finance, and Statistics. Each example consists of two parent-paper summaries paired with the core insight of a downstream paper, where the insight is a concise description of the paper's primary contribution automatically constructed by an LM from the downstream paper PDF. We also introduce an automatic evaluation method that uses an LM judge to give a \textbf{similarity score} comparing the model-generated insight to the ground-truth insight. Expert evaluation shows that these LM-judge scores are positively correlated with human ratings (Spearman's $\rho = 0.761$, $p < 0.001$).

We then introduce \ourmethod{}, a 4B-parameter language model trained via reinforcement learning (RL) to generate the core insight of a downstream paper from its parent papers. By fine-tuning with GRPO \citep{shao2024deepseekmath} to maximize the similarity scores on the research insights, the model learns to reason about connections between the input papers and generate insights that are more similar to those from real downstream papers (Figure~\ref{fig:giants_fig1}, bottom). This approach outperforms simply distilling the expert insight (even with reasoning traces).

We evaluate both proprietary and open LMs, including \gemtwopro{}, \gemthreepro{}, \qwenfourb{}, on \ourbenchmark{}. \qwenfourb{} performs similarly to \gemtwopro{} and \gemthreepro{}, despite being a smaller open model. Moreover, the similar performance of \gemtwopro{} and \gemthreepro{} suggests that frontier LMs are not simply getting better at insight anticipation via scaling. In contrast, our \ourmethod{} outperforms these baselines and generalizes zero-shot to unseen domains, achieving a 34\% improvement in similarity score over \gemthreepro{}. \ourmethod{} also produces insights that human evaluators judge to be more conceptually clear than those of the base model. In addition, SciJudge-30B~\citep{tong2026ai}, a third-party judge trained to compare research abstracts by likely citation impact, prefers \ourmethod{} over the base model in 68\% of pairwise comparisons. 

We summarize our main contributions as follows:
\begin{itemize}
    \item \textbf{Insight anticipation.} We introduce a new literature-grounded generation task that isolates the synthesis phase of scientific discovery by asking models to predict a downstream paper’s core insight from its parent papers.
    \item \textbf{\ourbenchmark{} and evaluation metric.} We construct a benchmark of 17k tuples of parent and downstream papers from arXiv across eight domains, together with an LM-based auto-evaluator for measuring the similarity between generated and ground-truth insights.
    \item \textbf{\ourmethod{}.} We train a model for insight anticipation via RL using similarity-based rewards and show that it outperforms a range of proprietary and open LMs such as \gemthreepro{}, improves conceptual clarity of insights, and generalizes zero-shot to unseen domains.
\end{itemize}

\section{Defining and Instantiating a Benchmark for Insight Anticipation}
\label{sec:dataset_construction}

\paragraph{Task Definition.} 
We define an \emph{insight} as a concise, natural language description of a paper's primary methodological or empirical advance. Building on this, we introduce the task of \emph{insight anticipation}: given an input context comprising the content of two parent papers, $x = (x_A, x_B)$, the goal is to generate the key insight of a downstream paper that builds on both parent papers ($A$ and $B$). We denote this downstream insight as the ground-truth insight, $y^*$, which emerges from the synthesis of the two foundational papers. While $y^*$ is just one of many possible subsequent ideas, it serves as a measurable proxy for the next conceptual leap. This setup allows us to probe a model's ability to reason about the joint influence of $A$ and $B$, challenging it to generate a research insight, $\hat{y}$, that is semantically similar to $y^*$. Figure~\ref{fig:insight_generation_prompt} shows the insight generation prompt used for training and evaluation of all models. To manage context constraints during training and inference, we intentionally restrict the context to two parent papers, establishing a conservative lower bound on the broader problem. Furthermore, we bypass parent discovery to focus entirely on whether meaningful insight synthesis is possible given fixed inputs. 

\paragraph{Dataset Construction.}
We create a dataset $\{((x_A, x_B)_i, y^*_i)\}_{i=1}^N$ as follows. We collect $17{,}839$ papers from arXiv that are published between May 23rd, 2007, and January 23rd, 2026, according to their latest update date on arXiv. Since arXiv is not peer reviewed, the raw corpus may contain noisy or non-substantive documents. To mitigate this, we retain only papers with at least two citations.\footnote{We use citation counts from Semantic Scholar as a proxy for paper quality.} We download these papers as a PDF. For each paper, we prompt \gemtwoflash{} to identify two prior papers that this paper explicitly cites and builds upon by combining their ideas in a synergistic way. We also ask \gemtwoflash{} to explain the synergy (see the full prompt in Figure~\ref{fig:upstream_citation_prompt}). We then download the two parent papers as a PDF and use their content as input context $x$. Ideally, we would like to take the full paper content as input to the model and ask it to generate a downstream insight. However, given the context-length limitations of existing LMs as well as the high inference cost, we instead opt to prompt \gemtwoflash{} to summarize the paper. We then use paper summaries as the input. 
We use the following prompt to summarize each paper into key insights and contributions: ``\textit{Summarize the document, clearly describing the method used and highlighting the key insights or findings. Provide sufficient detail so that the approach and main contributions are fully understood.}'' 
For $y^*$, we use the synergy explanation that is part of the output to the parent-identification prompt in Figure~\ref{fig:upstream_citation_prompt}. We cannot use the raw explanation directly as the target because it refers to a future downstream paper and talks about the insight in the context of the two parent papers. Instead, we would like the model to generate a standalone insight statement directly from the two parent papers alone. Therefore, we prompt \gemthreepro{} to rewrite the insight and explanation without reference to the downstream paper (see prompt in Appendix~\ref{appendix:rewriting_insight_prompt}). After constructing all $(x,y^*)$ pairs, for each unique pair of parent papers $x=(x_A, x_B)$, we keep the insight $y^*$ from the most cited downstream paper in order to prioritize generating more impactful insights.

\begin{wrapfigure}{r}{0.44\textwidth}
    \centering
    \vspace{-3mm}
    \includegraphics[width=0.6\linewidth]{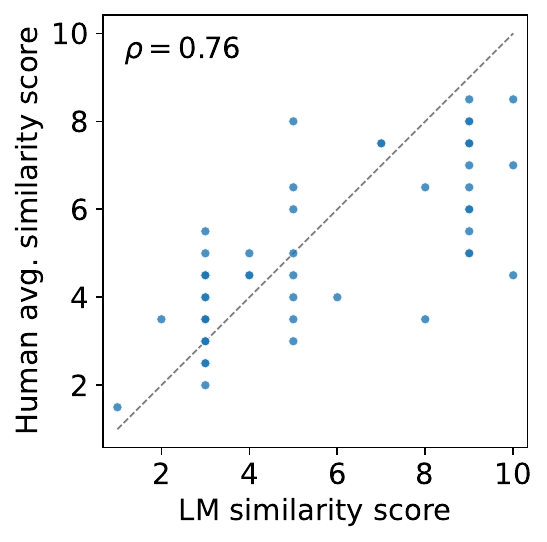}
    \caption{\footnotesize \textbf{LM judgements of insight similarity correlate with human judgements.} We ask both human annotators and the LM judge (\gemthreepro{}) to score the similarity between a model-generated insight and a ground-truth downstream insight on a 1–10 scale. The LM’s scores are positively correlated with average human ratings (Spearman $\rho = 0.761$, $p < 0.001$, $n = 60$).}
    \label{fig:lm_vs_human_scatterplot}
    \vspace{-0.5cm}
\end{wrapfigure}

\paragraph{Temporal Hold-Out Evaluation.}
To assess generalization, we conduct all evaluations on a \emph{future held-out test set} consisting of downstream papers published after the training cutoff date. We split the dataset by the publication date of the downstream paper in each $(x, y^*)$ pair. Papers published before July 1st, 2023 are used for training. To study the ability of models to generalize to new domains, we further restrict the training set to papers tagged with cs.CL (Computation and Language), yielding $N=10{,}335$ training examples. For evaluation, we consider papers published after July 1, 2023 and randomly sample up to $600$ papers from each domain (see the category taxonomy in Appendix~\ref{appendix:domain_mapping}), resulting in a test set of $7{,}504$ papers. Although we deduplicate the dataset by parent pair, some test examples may still share at most one parent paper with a training example. To address this, we also report results on a stricter subset of the test set, \textbf{Test-unseen-parents} ($N=5{,}294$), which excludes any test example that shares a parent with the training set.

\paragraph{Evaluation Metric.}
For each generation, we prompt an LM judge to give a \textbf{similarity score} ranging from 1 to 10 (higher is better) comparing the model-generated insight $\hat{y}$ to the ground-truth insight $y^*$ (see Figure~\ref{fig:similarity_judge_prompt} for the prompt). We use \gemthreepro{} as the primary judge model. To assess the reliability of the LM-based evaluation, we conduct a human evaluation study on $30$ pairs of insights generated by \qwenfourb{} and \ourmethod{} ($n = 60$). Two human annotators, who are PhD students in Computer Science, independently rate the similarity between model-generated insights and ground-truth insights using the same rating scale as the LM judge. The LM’s scores show a statistically significant positive correlation with the average scores across human annotators (Figure~\ref{fig:lm_vs_human_scatterplot}). In particular, we observe a Spearman rank correlation of $\rho = 0.761$ ($p < 0.001$). More details can be found in Appendix~\ref{app:human_eval_similarity}.

\paragraph{Conceptual View.} Conceptually, this framework can be viewed as an \emph{auto-encoding task}~\cite{kingma2022autoencodingvariationalbayes} over the citation graph. A target paper is passed through a highly lossy channel, the summaries of its two parent papers, from which the model must successfully reconstruct the original paper's core insight. By linearizing the citation graph into input-target pairs, we challenge the model to recreate the conceptual leap required to bridge adjacent nodes.

\begin{AIbox}{Takeaways of Insight Anticipation Instantiation}
We introduce the \emph{insight anticipation} task, which challenges models to generate a future research insight by synthesizing the summaries of two parent papers. To evaluate this, we construct a dataset of over $17k$ examples from arXiv with LM-extracted ground truths, using a temporal and cross-domain hold-out split and a human-validated LM judge for scoring.
\end{AIbox}
\section{Training an Insight Anticipation Model}
We explore two training paradigms for insight anticipation: (1) distillation via supervised fine-tuning (SFT), leveraging ground-truth insights and rationalization from target downstream papers, and (2) reinforcement learning (RL) via similarity optimization. Across all experiments, we use \qwenfourb{} as the base model. Next, we will detail the specific formulations of these training methodologies.

\subsection{Ground Truth Insight Distillation via Supervised Fine-Tuning (SFT)}
Our first approach is to fine-tune the base model to generate a downstream paper’s core insight from the summaries of its two parent papers. We investigate two distinct SFT strategies to achieve this. In our standard SFT approach, the model is directly fine-tuned to map the input context (summaries of parent papers), $x$, to the target ground-truth insight, $y^*$ (from the downstream paper). This process optimizes the standard cross-entropy loss for autoregressive LMs over the paired examples $(x, y^*)$.

To bridge the logical gap between the source papers and the final insight, we also evaluate an SFT strategy enhanced with chain-of-thought reasoning~\citep{wei2023chainofthoughtpromptingelicitsreasoning}, which we denote as \emph{SFT-think}. In this setup, we introduce an intermediate synthetic rationalization step, $z$. To generate a supervision target, we prompt a high-capacity teacher model (\gemthreepro{}) to generate a detailed chain-of-thought that logically deduces the ground-truth insight conditioned on the parent paper summaries (as detailed in Section~\ref{sec:dataset_construction}). The base model is then trained on augmented tuples $(x, z, y^*)$, learning to sequentially predict the rationale followed by the final insight. This approach explicitly encourages the model to internalize the step-by-step inferential process rather than merely memorizing the final output mapping. This follows distillation approaches in reasoning as seen in works such as OpenThoughts~\citep{guha2025openthoughtsdatarecipesreasoning} and s1~\citep{muennighoff2025s1simpletesttimescaling}.

\subsection{Reinforcement Learning for Insight Anticipation via Similarity Optimization}
So far, we have explored distilling insights via supervised fine-tuning (SFT). However, in complex tasks such as scientific discovery, the downstream paper's insights can be difficult to clone directly. This difficulty often arises when the downstream insight has low likelihood under the policy, or when the model lacks the capacity to adequately capture the distribution of the parent insight. Therefore, we explore an alternative approach: using semantic similarity to the downstream paper's insights $y^{*}$ as a proxy reward. Formally, given a ground-truth insight from the downstream paper $y^{*}$, we define the proxy reward for a predicted insight $\hat{y}$ as: 
\begin{align}
    r_{\text{sim}}(\hat{y}) = \textit{similarity}(\hat{y}, y^{*})
\end{align}
where $\textit{similarity}$ measures the semantic equivalence between the generated insight and downstream paper's insight measured using an LM-as-a-Judge~\citep{gu2025surveyllmasajudge}, matching the evaluation criterion in Section~\ref{sec:dataset_construction}. This proxy reward is then optimized via RL to recover the behavior of the ground-truth insight of the downstream paper conditioned on the two prespecified parent papers.

We optimize this reward using Group Relative Policy Optimization (GRPO) \citep{shao2024deepseekmath}. In particular, for each input context $x$, we sample a group of $G=8$ candidate insights from the current policy. An LM judge evaluates these candidates, and GRPO updates the policy relative to the sampled group (illustrated in Figure~\ref{fig:giants_fig1}, bottom). GRPO is well-suited for this setup because it avoids the need to train and maintain a separate, memory-intensive value function model (see implementation details in Appendix~\ref{app:hyperparameters}).

Crucially, to mitigate the risk of reward hacking and ensure a rigorous evaluation, we enforce a strict separation between the training and testing judges. We use \gemtwoflash{} as the active reward model during GRPO training, while reserving the independent \gemthreepro{} model exclusively for the final evaluation phase. This decoupling guarantees a more objective, unbiased assessment of the model's true generalization capabilities. We additionally evaluate with other LM judges (e.g., \qwenfourteenb{}) to showcase robustness across model families.

\begin{AIbox}{Takeaways of Training an Insight Anticipation Model}
We optimize our insight anticipation model using reinforcement learning to maximize semantic similarity to downstream ground-truth insights. We strictly separate the LM judges used for training and final evaluation to mitigate the risk of reward hacking.
\end{AIbox}

\section{Experimental Evaluation on \ourbenchmark{}}

Our experimental evaluation studies whether \ourmethod{} improves insight anticipation on \ourbenchmark{}. We compare \ourmethod{} against proprietary and open-weight language models, as well as supervised fine-tuning baselines built from the same base model. Unless otherwise noted, we evaluate all Qwen-based models (Base, SFT, SFT-think, and \ourmethod{}) using the recommended Qwen thinking-mode decoding settings: temperature $= 0.6$, \texttt{top-p} $= 0.95$, \texttt{top-k} $= 20$, and \texttt{min-p} $= 0$. For the Gemini baselines, we use temperature $= 0$. We report results on both the full test set and the stricter \textbf{Test-unseen-parents} subset, which excludes any test example that shares a parent paper with the training set. Unless otherwise noted, our primary evaluation uses \gemthreepro{} as the LM judge, and we additionally evaluate with other LM judges (e.g., \qwenfourteenb{}, \gemtwopro{}) to test whether model rankings are robust across judges.

Through these experiments, we address three questions: (1) how well current frontier and open-source language models perform on insight anticipation, (2) whether directly optimizing for insight similarity using RL (\ourmethod{}) improves insight anticipation, and (3) whether a model trained on one domain transfers to unseen domains.

\begin{figure}[h]
  \centering

  \includegraphics[width=0.85\linewidth]{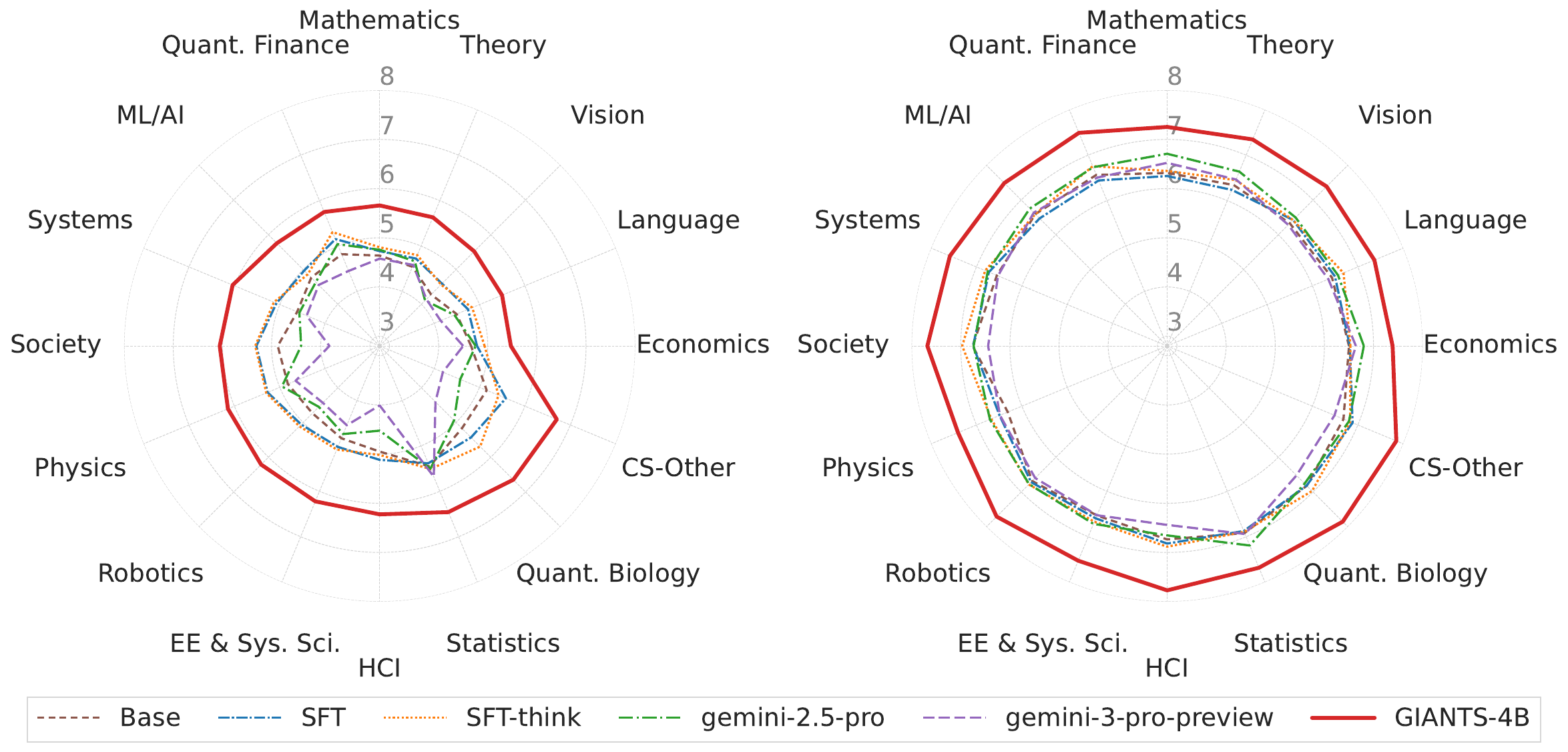}
  \vspace{-3mm}
  \captionof{figure}{\footnotesize \textbf{Similarity scores on \ourbenchmark{} (higher is better).} \ourmethod{} consistently achieves the highest similarity scores across domains. \textit{(left)} scores from our primary evaluation judge \gemthreepro{}. \textit{(right)} scores from \qwenfourteenb{}.}
  \label{fig:all_domain_radar_comparison_gemini-3-pro-preview_vs_Qwen3-14B}

  \vspace{4mm}

  \includegraphics[width=0.85\linewidth]{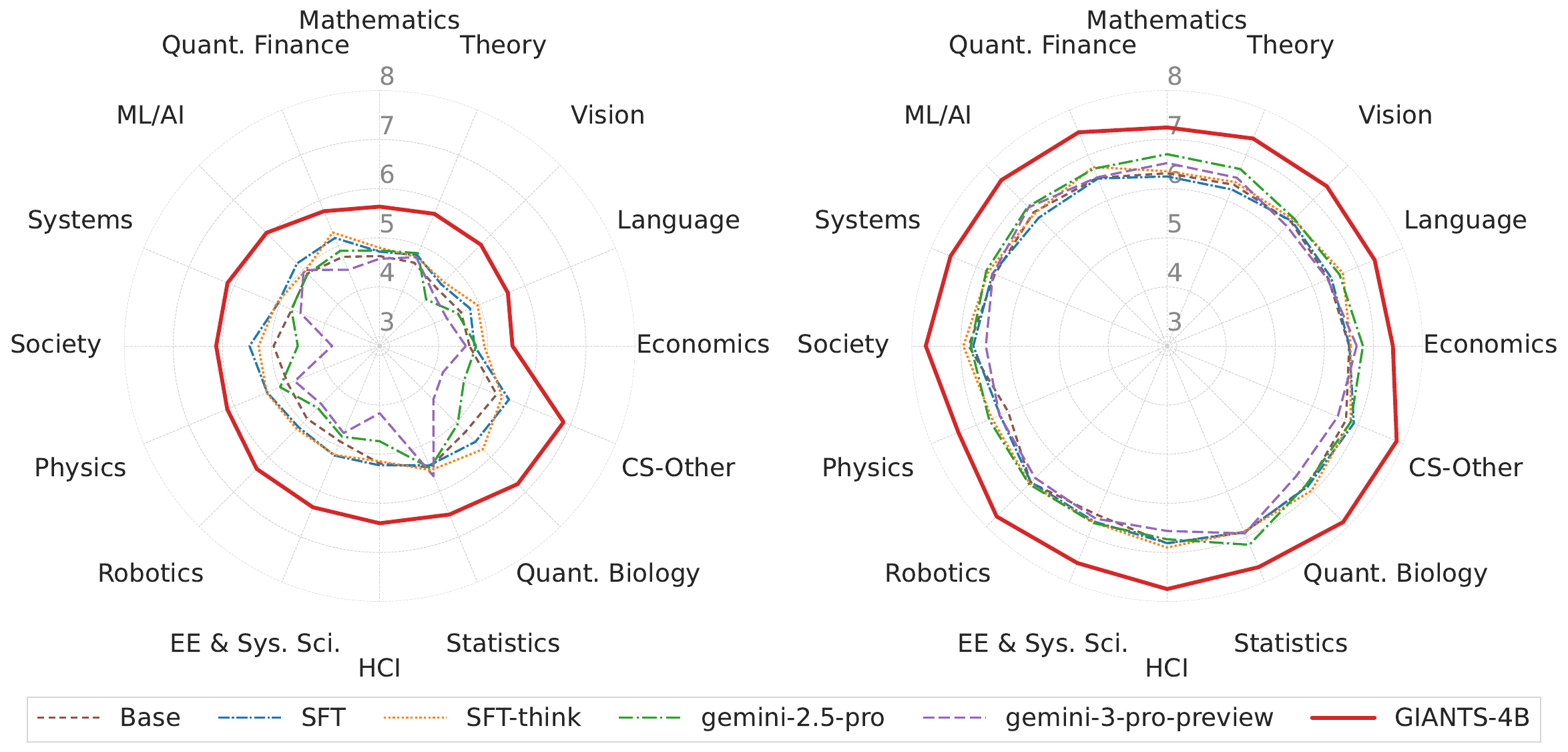}
  \vspace{-3mm}
  \captionof{figure}{\footnotesize \textbf{Similarity scores on a subset of \ourbenchmark{} containing only unseen parent papers.} \ourmethod{} generalizes to new domains and distribution of parent papers, as judged by \gemthreepro{} \textit{(left)} and \qwenfourteenb{} \textit{(right)}.}
  \label{fig:all_domain_radar_comparison_gemini-3-pro-preview_vs_Qwen3-14B_unseen_parents}
  \vspace{-3mm}
  
\end{figure}
\textbf{1) Insight anticipation is a challenging task even for large proprietary models.} Despite their success across diverse NLP benchmarks, current frontier models exhibit significant limitations in literature-grounded synthesis. As shown in Figure~\ref{fig:all_domain_radar_comparison_gemini-3-pro-preview_vs_Qwen3-14B}, the base open model, \qwenfourb{}, achieves an average similarity score of only $4.75$ (see raw statistics corresponding to Figure~\ref{fig:all_domain_radar_comparison_gemini-3-pro-preview_vs_Qwen3-14B} in Appendix~\ref{appendix:quantitative_analysis}). Based on our LM-based evaluation rubric (Figure~\ref{fig:similarity_judge_prompt}), a score in this range indicates that while generated insights may align topically with the ground-truth insights, they fail to capture the core scientific contribution or technical nuance. Notably, significantly larger proprietary models, including \gemtwopro{} and \gemthreepro{}, perform similarly to the smaller \qwenfourb{} baseline. This finding suggests that literature-grounded synthesis capabilities do not scale linearly with model size alone, reinforcing the necessity for specialized training paradigms like \ourmethod{}.

\begin{wrapfigure}{r}{0.4\textwidth}
    \centering
    \includegraphics[width=\linewidth]{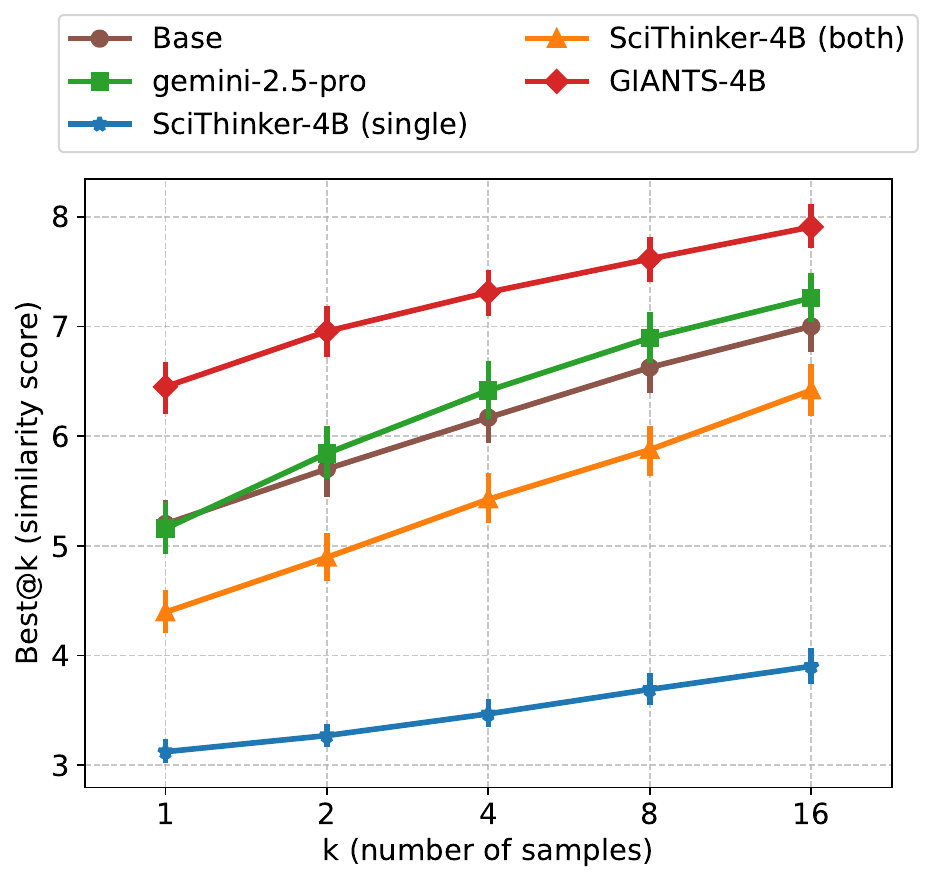}
    \vspace{-3mm}
    \caption{\footnotesize \textbf{\ourmethod{} remains strongest under test-time scaling.} As the number of samples per example increases, \ourmethod{} consistently outperforms the base model, \gemtwopro{}, and SciThinker-4B~\citep{tong2026ai}, which is a scientific-ideation model trained using a citation-preference reward model. Error bars show 95\% confidence intervals.}
    \vspace{-5mm}
    \label{fig:best_at_k}
\end{wrapfigure}
\textbf{2) Directly optimizing similarity scores substantially improves insight anticipation.} \ourmethod{} achieves the highest performance among all evaluated methods. Figure~\ref{fig:all_domain_radar_comparison_gemini-3-pro-preview_vs_Qwen3-14B} shows that, on the full future held-out test set, \ourmethod{} consistently outperforms the base model, both supervised baselines, and the proprietary baselines under two different judge LMs. Relative to \gemthreepro{}, \ourmethod{} achieves a 35\% improvement in similarity score on the full test set. Figure~\ref{fig:all_domain_radar_comparison_gemini-3-pro-preview_vs_Qwen3-14B_unseen_parents} shows that this advantage persists on the stricter \textbf{Test-unseen-parents} split, where GIANTS-4B achieves a 34\% improvement over \gemthreepro{} (see raw statistics corresponding to Figure~\ref{fig:all_domain_radar_comparison_gemini-3-pro-preview_vs_Qwen3-14B} and Figure~\ref{fig:all_domain_radar_comparison_gemini-3-pro-preview_vs_Qwen3-14B_unseen_parents} in Appendix~\ref{appendix:quantitative_analysis}). In contrast, standard SFT and SFT-think both improve slightly over the base model. These results suggest that standard imitation learning provides a modest benefit for this synthesis task; however, RL with a similarity-based reward effectively aligns the model's capabilities with target human insights. This performance trend remains robust as we increase the number of samples to optimize the similarity score via inference-time scaling, as illustrated in Figure~\ref{fig:best_at_k}. Since best-of-$k$ evaluation requires scoring many samples per example, we conduct the inference-time scaling evaluation in Figure~\ref{fig:best_at_k} on a subset of 480 examples sampled from \ourbenchmark{}, using \gemthreeflash{} as the LM judge for cost considerations. We use temperature $= 0.6$ for all models in this evaluation.

In Figure~\ref{fig:best_at_k}, we additionally compare against SciThinker-4B~\citep{tong2026ai}, a related scientific-ideation model rather than a direct same-task baseline. In the original setting of Tong et al.~\citep{tong2026ai}, SciThinker takes as input a single seed paper’s title and abstract and outputs a follow-up research idea. Its training objective is to generate ideas with high potential impact under a citation-preference reward model. By contrast, our task requires synthesizing two parent papers to generate the core insight of a downstream paper. To test whether this kind of general scientific ideation training transfers to insight anticipation, we evaluate two variants: SciThinker-4B (single), which is given one parent paper, and SciThinker-4B (both), which is given both parent papers. Figure~\ref{fig:best_at_k} shows that using both parents improves SciThinker-4B, but it still remains well below \ourmethod{} across all $k$. This suggests that general scientific ideation training may not transfer well to insight anticipation, and that strong performance on our task requires training aligned with literature-grounded multi-paper synthesis rather than open-ended follow-up ideation.

\begin{figure}[t]
  \centering
  \includegraphics[width=0.9\linewidth]{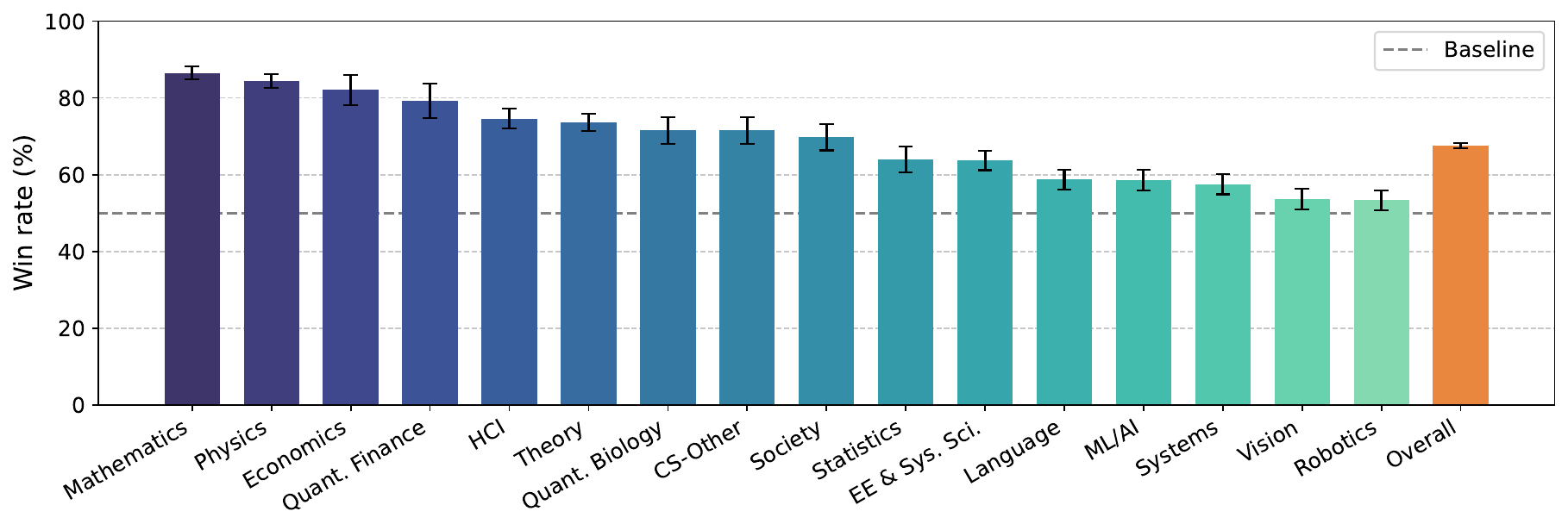}
  \vspace{-3mm}
  \caption{\footnotesize \textbf{\ourmethod{} achieves a 68\% overall win rate against the base model under SciJudge-30B.} We evaluate pairwise preferences using SciJudge-30B \citep{tong2026ai}, a third-party judge trained to compare research abstracts by likely citation impact. This provides complementary evidence that optimizing for insight anticipation also improves performance under an independent, impact-oriented evaluation signal. Error bars show standard errors.}
  \label{fig:winrate_taste}
  \vspace{2mm}
  \begin{minipage}[t]{0.449\textwidth}
    \centering
    \includegraphics[width=0.82\linewidth]{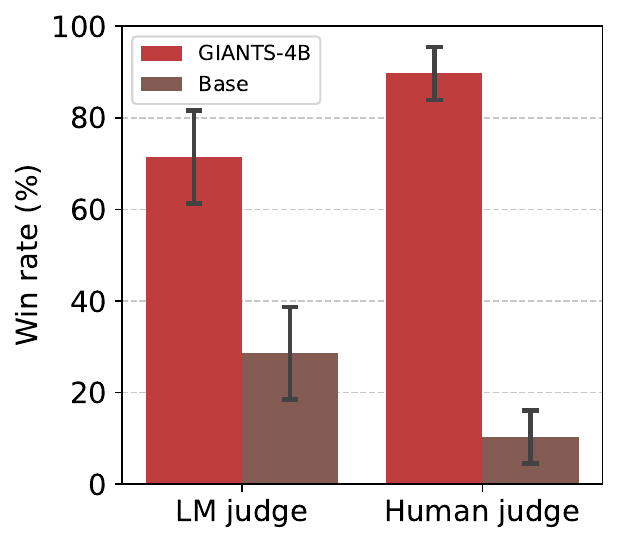}
    \caption{\footnotesize \textbf{Win rates of \ourmethod{} against the base model for insight similarity.} We ask both human annotators and an LM judge to score the similarity between a model-generated insight and a ground-truth downstream insight and find that \ourmethod{} better matches the ground-truth insight. Error bars show standard errors.}
    \label{fig:win_rate_human_eval_similarity}
    
  \end{minipage}
  \hfill
  \begin{minipage}[t]{0.473\textwidth}
    \centering

    \includegraphics[width=0.76\linewidth]{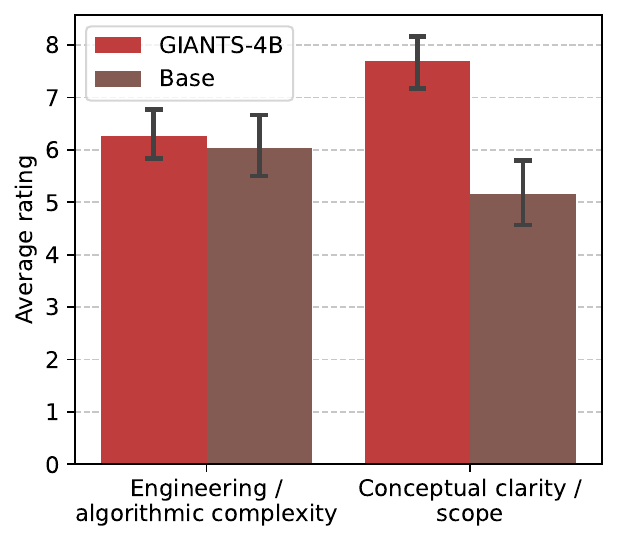}
    \caption{\footnotesize \textbf{\ourmethod{} produces insights with similar perceived algorithmic complexity but substantially higher conceptual clarity compared to the base model.} Human annotators assess the feasibility of generated insights along two axes: (1) engineering / algorithmic complexity and (2) conceptual clarity. Error bars show 95\% confidence intervals.}
    \label{fig:feasibility_bar_plot}
  \end{minipage}
  \vspace{-3mm}
\end{figure}

To further validate these gains, we also evaluate generations using SciJudge-30B, a third-party model from \citet{tong2026ai} trained to compare research abstracts by likely citation impact. Under this external preference signal, \ourmethod{} achieves a 68\% overall win rate against the base model (Figure~\ref{fig:winrate_taste}), with variation across domains. While this metric is only a proxy for research impact, it provides complementary evidence that optimizing for similarity produces outputs that are also preferred by an independent quality-oriented judge. To reduce order effects, we evaluate each pair twice with reversed presentation order and filter inconsistent judgments.

We supplemented these automated evaluations with two independent preliminary human studies. First, we ask two human annotators to rate the similarity between model-generated insights and their corresponding ground-truth insights for 30 head-to-head pairs from the base model and \ourmethod{}, using the same rating scale as the LM judge (Figure~\ref{fig:similarity_judge_prompt}). We then compare the win rate of \ourmethod{} against the base model in terms of alignment with the ground-truth insight (Figure~\ref{fig:win_rate_human_eval_similarity}). In this comparison, \ourmethod{} achieves a $71.4\%$ win rate under the LM judge (\gemthreepro{}) and an $89.7\%$ win rate under human evaluation. Second, we assess the feasibility of generated insights along two axes: algorithmic complexity and conceptual clarity. Figure~\ref{fig:feasibility_bar_plot} shows that while \ourmethod{} produces insights of similar algorithmic complexity to the base model, it significantly improves conceptual clarity, making the generated ideas more interpretable and actionable. Further details regarding the human study methodology are provided in Appendix~\ref{app:human_eval}.

\textbf{3) \ourmethod{} zero-shot generalizes to new domains and distribution of parent papers}. 

Crucially, the synthesis capabilities acquired by \ourmethod{} are not restricted to its training data. Although trained exclusively on papers from the Language domain (CS.CL), Figure~\ref{fig:all_domain_radar_comparison_gemini-3-pro-preview_vs_Qwen3-14B} shows consistent gains across all evaluated domains. Moreover, these evaluations are conducted on papers published after the training cutoff, so the gains reflect performance on temporally held-out downstream literature rather than memorization of the training set. This suggests that the model learns a generalizable mechanism for combining disparate ideas rather than memorizing domain-specific heuristics. In addition, Figure~\ref{fig:all_domain_radar_comparison_gemini-3-pro-preview_vs_Qwen3-14B_unseen_parents} shows model performance on a subset of test examples whose parent papers are entirely unseen during training. \ourmethod{} remains the top-performing method on this subset, suggesting that its gains are not driven by partial overlap in parent-paper lineage.

\textbf{Evaluation reliability check}. 
To ensure our findings are robust and that similarity scores are not biased toward the specific LM judge that was used during training, we conducted a cross-model validation using an independent LM judge, \qwenfourteenb{} (Figure~\ref{fig:all_domain_radar_comparison_gemini-3-pro-preview_vs_Qwen3-14B}, right). This secondary evaluation corroborated our primary findings, ranking \ourmethod{} as the top-performing model across all metrics, albeit with a slightly more compressed performance margin. We also report results under three additional judge models in Appendix~\ref{appendix:ablations_with_diverse_lm_judges}, with consistent findings across model sizes.

\begin{figure}[t]
    \centering
    \includegraphics[width=\linewidth]{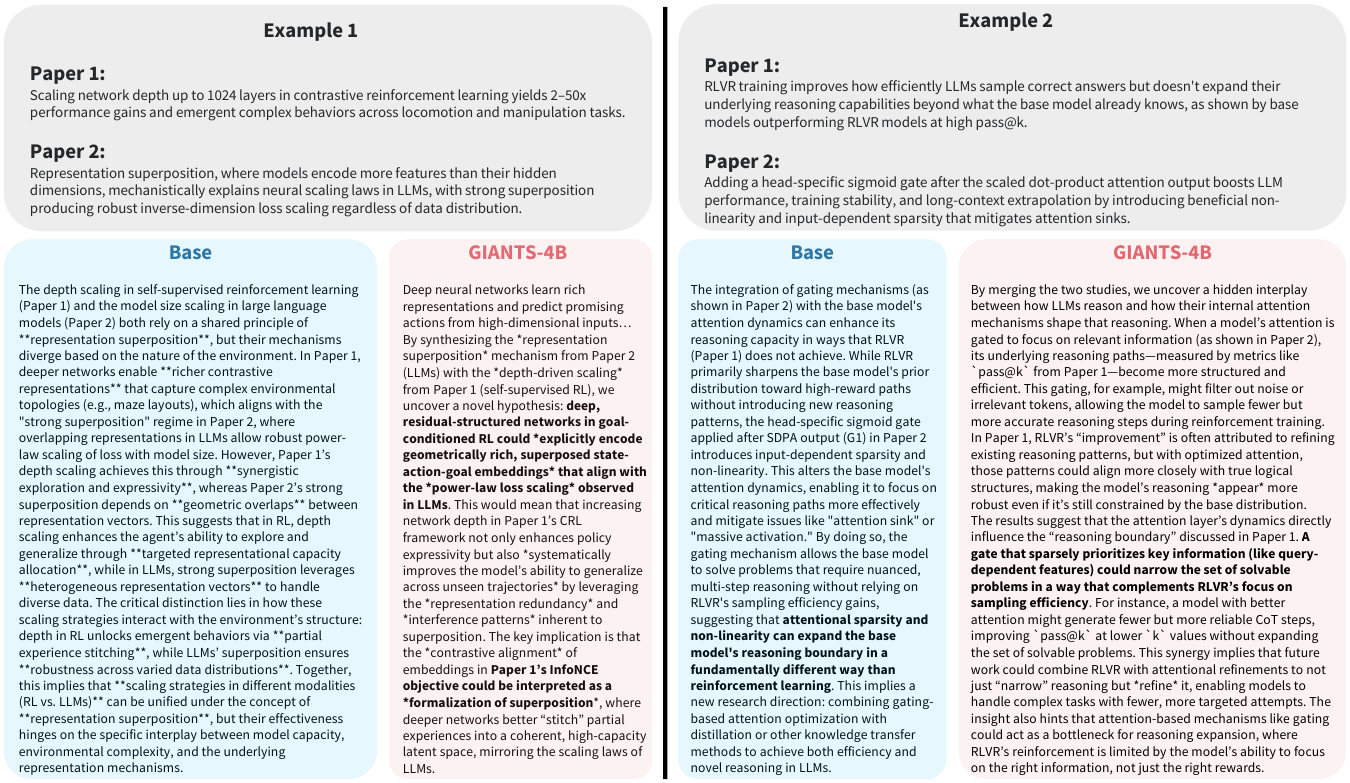}
    \vspace{-3mm}
    \caption{\footnotesize \textbf{Qualitative comparison of insights derived from NeurIPS 2025 award-winning papers.} \textit{(left)} \ourmethod{} identifies a more concrete cross-paper mechanism than the base model (\qwenfourb{}). \textit{(right)} \ourmethod{} produces a grounded, more plausible interaction, while the base model makes a broader conjecture that is less directly grounded in the parent papers. We show a one-sentence summary of each parent paper for readability. These are illustrative abbreviations only and are not the full parent-paper summaries used as model input.}
    \label{fig:qualitative}
    \vspace{-3mm}
\end{figure}
\textbf{Qualitative Comparison of Models.}
We present qualitative examples to demonstrate that our generated insights meaningfully integrate concepts from prior literature. We use NeurIPS 2025 award-winning papers as representative high-quality parent papers and compare the insights produced by \ourmethod{} against those generated by the base model given these parent papers.

The first example (Figure~\ref{fig:qualitative}, left) shows insights derived from \citet{wang20251000} and \citet{liu2025superposition}. While the base model merely summarizes the parent papers without synthesizing a novel perspective, \ourmethod{} proposes a concrete mechanistic connection between the two studies. The second example (Figure~\ref{fig:qualitative}, right) shows insights based on \citet{yue2025does} and \citet{qiu2025gated}. In this example, the base model generates an overly ambitious claim, suggesting that gating mechanisms expand reasoning boundaries in ways that RLVR cannot, which is unsupported by the source texts. In contrast, \ourmethod{} remains grounded in the actual findings while still identifying a non-trivial connection: that attention gating may dictate how effectively reinforcement learning concentrates probability mass across useful reasoning trajectories. This comparison underscores a critical distinction between boldness and genuine insightfulness. While the base model's novelty relies on unsupported extrapolation, \ourmethod{} maintains a narrower scope to identify a highly plausible interaction between the foundational works.

These qualitative examples are consistent with the quantitative results and human evaluations. \ourmethod{} more often produces insights that connect the parent papers in a concrete, interpretable, and grounded way, rather than simply restating their contents or making broader but weakly supported claims.

\begin{AIbox}{Takeaways of Experimental Evaluation}
\ourmethod{} significantly outperforms both frontier models and SFT models on our insight anticipation task. \ourmethod{} produces insights that are rated by human experts as more conceptually clear than those of the base model. A third-party citation-preference judge also prefers \ourmethod{}'s outputs over the base model. Furthermore, despite training exclusively on a single domain, the model successfully zero-shot generalizes its synthesis capabilities across diverse, unseen scientific disciplines and temporally held-out literature.
\end{AIbox}

\section{Related Work}
\paragraph{AI for Research.}
Many works have studied how to use LMs for different components of the scientific research pipeline, including literature search~\citep{openai_deep_research, zheng-etal-2025-deepresearcher}, idea generation~\citep{si2024can, wang-etal-2024-scimon, li2024chain, zhao2025ramon, tong2026ai}, idea execution~\citep{si2025ideation, si2026towards, karpathy2026autoresearch}, and paper review~\citep{weng2024cycleresearcher}. \citet{asai2026synthesizing} evaluate the ability of models to answer literature synthesis questions by identifying relevant papers and generating long-form responses with citations. Many existing methods for idea generation either directly prompt LMs to perform open-ended brainstorming or expose LMs to a set of prior works without requiring cross-paper synthesis~\citep{si2024can, li2024chain}. \textsc{SciMON}~\citep{wang-etal-2024-scimon} studies literature-grounded scientific idea generation, but its setting is different from ours: \textsc{SciMON} takes background problem contexts as input, retrieves literature inspirations, and generates open-ended ideas optimized for novelty relative to prior work. By contrast, we focus on synthesizing insights from prior work and do not optimize for open-ended novelty. A concurrent work trains models to judge research impact from citation signals, then uses the learned judge as a reward model for idea generation~\citep{tong2026ai}. In contrast, we focus on the problem of synthesizing insights from two parent papers, using similarity to ground-truth downstream insights as the training signal. A related line of work studies scientific progress as a forecasting problem. \textsc{PreScience} decomposes the research process into collaborator prediction, prior work selection, contribution generation, and impact prediction \citep{ajith2026prescience}. In its contribution generation task, models generate a future paper's title and abstract conditioned on prior work and other historical context, and are evaluated using an LM-based similarity metric. Compared to this broader scientific forecasting setup, we isolate the idea-synthesis problem and test whether models can generate the core insight of a downstream paper from a given pair of parent papers.

\paragraph{Literature-based Discovery.}
Literature-Based Discovery (LBD) studies how computational methods can uncover hidden links between seemingly unrelated bodies of research to infer novel and potentially useful knowledge \citep{swanson1986fish, swanson2008literature}. A classic example is Swanson’s hypothesis that fish oil could treat Raynaud’s syndrome: one set of papers suggested that fish oil reduces blood viscosity, while another linked high blood viscosity to Raynaud’s syndrome \citep{swanson1986undiscovered}. However, existing LBD approaches often struggle to scale to the volume of modern scientific literature, rely on domain-specific knowledge sources, or output ranked candidate connections rather than clear, standalone hypotheses that researchers can directly act on \citep{sebastian2017emerging, ganiz2005recent, bekhuis2006conceptual, smalheiser2012literature}. Our work is complementary but more focused: rather than identifying latent cross-literature links, we study whether a model can synthesize two parent papers into the core insight of a real downstream paper.
\section{Discussion, Limitations, Future Work}
In this work, we introduced \emph{insight anticipation} as a measurable paradigm for automated scientific discovery. By isolating the insight generation phase, we demonstrated that models can effectively synthesize the core insight of a downstream paper when provided with its foundational parent papers. Our results with \ourmethod{} indicate that the trajectory of scientific intuition is partially predictable, and that optimizing language models via reinforcement learning with similarity-based rewards is a highly effective training strategy for this task. However, this foundational formulation leaves several important limitations and open questions.

One limitation of the work is that we assume that downstream contributions derive from two parent papers due to context constraints, despite research ideas being heavily shaped by broader intellectual contexts. Furthermore, parent identification is imperfect. Citations do not always reflect true conceptual influence, and influential ideas may remain uncited. Finally, by assuming an oracle literature selection criterion, we explicitly decoupled insight generation from parent selection, which may not be feasible for some scientific tasks.

Future research can address the parent selection problem directly or integrate automated retrieval systems to unify parent selection and synthesis within a single end-to-end framework. Additionally, extending the task to accommodate multi-source lineage and developing evaluation metrics that prioritize conceptual novelty over textual similarity could be interesting. Ultimately, testing these models in active, human-in-the-loop research settings will determine their true potential as catalysts for scientific discovery.
\vspace{-0.15cm}
\section{Ethics Statement}
\vspace{-0.15cm}
This research involves training the \ourmethod{} model on a dataset of 17,839 publicly available arXiv pre-prints to synthesize scientific insights. While automating scientific ideation offers significant potential for discovery, it raises important ethical considerations regarding the proper attribution of foundational ideas, especially since academic citation graphs can be noisy and may not always reflect true conceptual influence. Additionally, there is an inherent risk of language models generating plausible but unverified scientific claims; our framework mitigates this by using reinforcement learning to directly align model outputs with grounded, human-derived insights. Finally, we acknowledge the environmental and computational costs associated with training models on high-performance GPUs, though utilizing a more efficient 4B parameter architecture helps limit this footprint compared to massive proprietary models.

\vspace{-0.15cm}
\section{Reproducibility Statement}
\vspace{-0.15cm}
To ensure the reproducibility of our results, we provide a comprehensive account of our methodology, code, and data. The source code for our models and experiments is available at the following repository: \url{https://github.com/joyheyueya/giants}. Our benchmark and model weights can be found in the following HuggingFace repository: \url{https://huggingface.co/giants2026}. Our implementation is built upon the verl framework (\url{https://verl.readthedocs.io/en/latest/}). All experimental details, including hyperparameter settings, are documented in Appendix~\ref{app:hyperparameters}. The computational experiments were conducted on a machine with NVIDIA A100 GPUs, and the required software dependencies are listed in the requirements.txt file within our code repository.

\section{Acknowledgments}
We thank Shirley Wu, Chenglei Si, Ryan Louie, Rui Li, Yoonho Lee, Jack Bai, Aviral Kumar, Andreas Stuhlmüller and others in the Stanford CoCoLab, Brunskill lab, and Iris Lab for discussions and feedback. This work was supported by the Junglee Corporation Stanford Graduate Fellowship. AS gratefully acknowledges the support of the NSF Graduate Research Fellowship Program, Modal Academic Research Program, and the Toyota Research Institute. CF was supported by Schmidt Sciences.

\clearpage

\bibliography{main}

@article{swanson2025virtual,
  title={The Virtual Lab of AI agents designs new SARS-CoV-2 nanobodies},
  author={Swanson, Kyle and Wu, Wesley and Bulaong, Nash L and Pak, John E and Zou, James},
  journal={Nature},
  pages={1--3},
  year={2025},
  publisher={Nature Publishing Group UK London}
}

@article{si2024can,
  title={Can llms generate novel research ideas? a large-scale human study with 100+ nlp researchers},
  author={Si, Chenglei and Yang, Diyi and Hashimoto, Tatsunori},
  journal={arXiv preprint arXiv:2409.04109},
  year={2024}
}

@article{si2025ideation,
  title={The Ideation-Execution Gap: Execution Outcomes of LLM-Generated versus Human Research Ideas},
  author={Si, Chenglei and Hashimoto, Tatsunori and Yang, Diyi},
  journal={arXiv preprint arXiv:2506.20803},
  year={2025}
}

@article{smalheiser2012literature,
  title={Literature-based discovery: Beyond the ABCs},
  author={Smalheiser, Neil R},
  journal={Journal of the American Society for Information Science and Technology},
  volume={63},
  number={2},
  pages={218--224},
  year={2012},
  publisher={Wiley Online Library}
}

@article{bekhuis2006conceptual,
  title={Conceptual biology, hypothesis discovery, and text mining: Swanson's legacy},
  author={Bekhuis, Tanja},
  journal={Biomedical digital libraries},
  volume={3},
  number={1},
  pages={2},
  year={2006},
  publisher={Springer}
}

@article{ganiz2005recent,
  title={Recent advances in literature based discovery},
  author={Ganiz, Murat C and Pottenger, William M and Janneck, Christopher D},
  journal={Journal of the American Society for Information Science and Technology, JASIST (Submitted)},
  year={2005}
}

@article{shao2024deepseekmath,
  title={Deepseekmath: Pushing the limits of mathematical reasoning in open language models},
  author={Shao, Zhihong and Wang, Peiyi and Zhu, Qihao and Xu, Runxin and Song, Junxiao and Bi, Xiao and Zhang, Haowei and Zhang, Mingchuan and Li, YK and Wu, Yang and others},
  journal={arXiv preprint arXiv:2402.03300},
  year={2024}
}

@article{sebastian2017emerging,
  title={Emerging approaches in literature-based discovery: techniques and performance review},
  author={Sebastian, Yakub and Siew, Eu-Gene and Orimaye, Sylvester O},
  journal={The Knowledge Engineering Review},
  volume={32},
  pages={e12},
  year={2017},
  publisher={Cambridge University Press}
}

@article{swanson1986fish,
  title={Fish oil, Raynaud's syndrome, and undiscovered public knowledge},
  author={Swanson, Don R},
  journal={Perspectives in biology and medicine},
  volume={30},
  number={1},
  pages={7--18},
  year={1986},
  publisher={Johns Hopkins University Press}
}

@incollection{swanson2008literature,
  title={Literature-based discovery? The very idea},
  author={Swanson, DR},
  booktitle={Literature-based discovery},
  pages={3--11},
  year={2008},
  publisher={Springer}
}

@article{asai2026synthesizing,
  title={Synthesizing scientific literature with retrieval-augmented language models},
  author={Asai, Akari and He, Jacqueline and Shao, Rulin and Shi, Weijia and Singh, Amanpreet and Chang, Joseph Chee and Lo, Kyle and Soldaini, Luca and Feldman, Sergey and D’Arcy, Mike and others},
  journal={Nature},
  pages={1--7},
  year={2026},
  publisher={Nature Publishing Group UK London}
}

@article{swanson1986undiscovered,
  title={Undiscovered public knowledge},
  author={Swanson, Don R},
  journal={The Library Quarterly},
  volume={56},
  number={2},
  pages={103--118},
  year={1986},
  publisher={University of Chicago Press}
}

@article{tong2026ai,
  title={AI Can Learn Scientific Taste},
  author={Tong, Jingqi and Li, Mingzhe and Li, Hangcheng and Yang, Yongzhuo and Mou, Yurong and Ma, Weijie and Xi, Zhiheng and Chen, Hongji and Liu, Xiaoran and Cheng, Qinyuan and others},
  journal={arXiv preprint arXiv:2603.14473},
  year={2026}
}

@misc{openai_deep_research,
    title={Introducing Deep Research},
    author={{OpenAI}},
    year={2025},
    howpublished={\url{https://openai.com/index/introducing-deep-research/}},
    note={Accessed: 2025-02-02}
}

@inproceedings{zheng-etal-2025-deepresearcher,
    title = "{D}eep{R}esearcher: Scaling Deep Research via Reinforcement Learning in Real-world Environments",
    author = "Zheng, Yuxiang  and
      Fu, Dayuan  and
      Hu, Xiangkun  and
      Cai, Xiaojie  and
      Ye, Lyumanshan  and
      Lu, Pengrui  and
      Liu, Pengfei",
    editor = "Christodoulopoulos, Christos  and
      Chakraborty, Tanmoy  and
      Rose, Carolyn  and
      Peng, Violet",
    booktitle = "Proceedings of the 2025 Conference on Empirical Methods in Natural Language Processing",
    month = nov,
    year = "2025",
    address = "Suzhou, China",
    publisher = "Association for Computational Linguistics",
    url = "https://aclanthology.org/2025.emnlp-main.22/",
    doi = "10.18653/v1/2025.emnlp-main.22",
    pages = "414--431",
    ISBN = "979-8-89176-332-6"
}

@inproceedings{wang-etal-2024-scimon,
    title = "{S}ci{MON}: Scientific Inspiration Machines Optimized for Novelty",
    author = "Wang, Qingyun  and
      Downey, Doug  and
      Ji, Heng  and
      Hope, Tom",
    editor = "Ku, Lun-Wei  and
      Martins, Andre  and
      Srikumar, Vivek",
    booktitle = "Proceedings of the 62nd Annual Meeting of the Association for Computational Linguistics (Volume 1: Long Papers)",
    month = aug,
    year = "2024",
    address = "Bangkok, Thailand",
    publisher = "Association for Computational Linguistics",
    url = "https://aclanthology.org/2024.acl-long.18/",
    doi = "10.18653/v1/2024.acl-long.18",
    pages = "279--299"
}

@article{si2026towards,
  title={Towards Execution-Grounded Automated AI Research},
  author={Si, Chenglei and Yang, Zitong and Choi, Yejin and Cand{\`e}s, Emmanuel and Yang, Diyi and Hashimoto, Tatsunori},
  journal={arXiv preprint arXiv:2601.14525},
  year={2026}
}

@article{ajith2026prescience,
  title={PreScience: A Benchmark for Forecasting Scientific Contributions},
  author={Ajith, Anirudh and Singh, Amanpreet and DeYoung, Jay and Kunievsky, Nadav and Kozlowski, Austin C and Tafjord, Oyvind and Evans, James and Weld, Daniel S and Hope, Tom and Downey, Doug},
  journal={arXiv preprint arXiv:2602.20459},
  year={2026}
}

@article{wang20251000,
  title={1000 layer networks for self-supervised rl: Scaling depth can enable new goal-reaching capabilities},
  author={Wang, Kevin and Javali, Ishaan and Bortkiewicz, Micha{\'L} and Eysenbach, Benjamin and others},
  journal={arXiv preprint arXiv:2503.14858},
  year={2025}
}

@article{liu2025superposition,
  title={Superposition yields robust neural scaling},
  author={Liu, Yizhou and Liu, Ziming and Gore, Jeff},
  journal={arXiv preprint arXiv:2505.10465},
  year={2025}
}

@article{yue2025does,
  title={Does reinforcement learning really incentivize reasoning capacity in llms beyond the base model?},
  author={Yue, Yang and Chen, Zhiqi and Lu, Rui and Zhao, Andrew and Wang, Zhaokai and Song, Shiji and Huang, Gao},
  journal={arXiv preprint arXiv:2504.13837},
  year={2025}
}

@article{qiu2025gated,
  title={Gated attention for large language models: Non-linearity, sparsity, and attention-sink-free},
  author={Qiu, Zihan and Wang, Zekun and Zheng, Bo and Huang, Zeyu and Wen, Kaiyue and Yang, Songlin and Men, Rui and Yu, Le and Huang, Fei and Huang, Suozhi and others},
  journal={arXiv preprint arXiv:2505.06708},
  year={2025}
}

@misc{karpathy2026autoresearch,
  author = {Karpathy, Andrej},
  title = {autoresearch: AI agents running research on single-GPU nanochat training automatically},
  year = {2026},
  publisher = {GitHub},
  journal = {GitHub repository},
  howpublished = {\url{https://github.com/karpathy/autoresearch}},
}

@misc{gu2025surveyllmasajudge,
      title={A Survey on LLM-as-a-Judge}, 
      author={Jiawei Gu and Xuhui Jiang and Zhichao Shi and Hexiang Tan and Xuehao Zhai and Chengjin Xu and Wei Li and Yinghan Shen and Shengjie Ma and Honghao Liu and Saizhuo Wang and Kun Zhang and Yuanzhuo Wang and Wen Gao and Lionel Ni and Jian Guo},
      year={2025},
      eprint={2411.15594},
      archivePrefix={arXiv},
      primaryClass={cs.CL},
      url={https://arxiv.org/abs/2411.15594}, 
}

@misc{wei2023chainofthoughtpromptingelicitsreasoning,
      title={Chain-of-Thought Prompting Elicits Reasoning in Large Language Models}, 
      author={Jason Wei and Xuezhi Wang and Dale Schuurmans and Maarten Bosma and Brian Ichter and Fei Xia and Ed Chi and Quoc Le and Denny Zhou},
      year={2023},
      eprint={2201.11903},
      archivePrefix={arXiv},
      primaryClass={cs.CL},
      url={https://arxiv.org/abs/2201.11903}, 
}

@misc{guha2025openthoughtsdatarecipesreasoning,
      title={OpenThoughts: Data Recipes for Reasoning Models}, 
      author={Etash Guha and Ryan Marten and Sedrick Keh and Negin Raoof and Georgios Smyrnis and Hritik Bansal and Marianna Nezhurina and Jean Mercat and Trung Vu and Zayne Sprague and Ashima Suvarna and Benjamin Feuer and Liangyu Chen and Zaid Khan and Eric Frankel and Sachin Grover and Caroline Choi and Niklas Muennighoff and Shiye Su and Wanjia Zhao and John Yang and Shreyas Pimpalgaonkar and Kartik Sharma and Charlie Cheng-Jie Ji and Yichuan Deng and Sarah Pratt and Vivek Ramanujan and Jon Saad-Falcon and Jeffrey Li and Achal Dave and Alon Albalak and Kushal Arora and Blake Wulfe and Chinmay Hegde and Greg Durrett and Sewoong Oh and Mohit Bansal and Saadia Gabriel and Aditya Grover and Kai-Wei Chang and Vaishaal Shankar and Aaron Gokaslan and Mike A. Merrill and Tatsunori Hashimoto and Yejin Choi and Jenia Jitsev and Reinhard Heckel and Maheswaran Sathiamoorthy and Alexandros G. Dimakis and Ludwig Schmidt},
      year={2025},
      eprint={2506.04178},
      archivePrefix={arXiv},
      primaryClass={cs.LG},
      url={https://arxiv.org/abs/2506.04178}, 
}

@misc{muennighoff2025s1simpletesttimescaling,
      title={s1: Simple test-time scaling}, 
      author={Niklas Muennighoff and Zitong Yang and Weijia Shi and Xiang Lisa Li and Li Fei-Fei and Hannaneh Hajishirzi and Luke Zettlemoyer and Percy Liang and Emmanuel Candès and Tatsunori Hashimoto},
      year={2025},
      eprint={2501.19393},
      archivePrefix={arXiv},
      primaryClass={cs.CL},
      url={https://arxiv.org/abs/2501.19393}, 
}

@misc{wadhwa2026createtestingllmsassociative,
      title={CREATE: Testing LLMs for Associative Creativity}, 
      author={Manya Wadhwa and Tiasa Singha Roy and Harvey Lederman and Junyi Jessy Li and Greg Durrett},
      year={2026},
      eprint={2603.09970},
      archivePrefix={arXiv},
      primaryClass={cs.CL},
      url={https://arxiv.org/abs/2603.09970}, 
}

@misc{kingma2022autoencodingvariationalbayes,
      title={Auto-Encoding Variational Bayes}, 
      author={Diederik P Kingma and Max Welling},
      year={2022},
      eprint={1312.6114},
      archivePrefix={arXiv},
      primaryClass={stat.ML},
      url={https://arxiv.org/abs/1312.6114}, 
}

@article{li2024chain,
  title={Chain of ideas: Revolutionizing research via novel idea development with llm agents},
  author={Li, Long and Xu, Weiwen and Guo, Jiayan and Zhao, Ruochen and Li, Xingxuan and Yuan, Yuqian and Zhang, Boqiang and Jiang, Yuming and Xin, Yifei and Dang, Ronghao and others},
  journal={arXiv preprint arXiv:2410.13185},
  year={2024}
}

@article{weng2024cycleresearcher,
  title={Cycleresearcher: Improving automated research via automated review},
  author={Weng, Yixuan and Zhu, Minjun and Bao, Guangsheng and Zhang, Hongbo and Wang, Jindong and Zhang, Yue and Yang, Linyi},
  journal={arXiv preprint arXiv:2411.00816},
  year={2024}
}

@article{zhao2025ramon,
  title={The Ramon Llull's Thinking Machine for Automated Ideation},
  author={Zhao, Xinran and Zheng, Boyuan and Si, Chenglei and Yu, Haofei and Liu, Ken and Zhou, Runlong and Li, Ruochen and Chen, Tong and Li, Xiang and Zhang, Yiming and others},
  journal={arXiv preprint arXiv:2508.19200},
  year={2025}
}

\newpage
\appendix
\onecolumn

\section{arXiv Paper Domain Classification}
\label{appendix:domain_mapping}
We use the \href{https://arxiv.org/category_taxonomy}{arXiv category taxonomy}. Since Computer Science has many subfields, we further group the arXiv category labels into broader macro-domains.
\begin{table}[htbp]
\centering
\resizebox{\textwidth}{!}{%
\begin{tabular}{ll}
\toprule
\textbf{Domain} & \textbf{arXiv Labels} \\
\midrule
Language & cs.CL \\
ML/AI & cs.LG, cs.AI, cs.NE, cs.MA \\
Robotics & cs.RO \\
Vision & cs.CV, cs.GR, cs.MM, cs.SD \\
Theory & cs.CC, cs.DS, cs.FL, cs.LO, cs.DM, cs.CG, cs.GT, cs.CR, cs.IT \\
Systems & cs.AR, cs.OS, cs.DC, cs.NI, cs.PF, cs.SY, cs.PL, cs.SE, cs.DB, cs.IR, cs.SI \\
Society & cs.CY \\
HCI & cs.HC \\
CS-Other & cs.ET, cs.GL, cs.OH, cs.DL, cs.NA, cs.MS, cs.CE, cs.SC \\
Economics & econ \\
Electrical Engineering \& Systems Science (EE \& Sys. Sci.) & eess \\
Mathematics & math \\
Physics & astro-ph, cond-mat, gr-qc, hep-ex, hep-lat, hep-ph, hep-th, math-ph, nlin, nucl-ex, nucl-th, physics, quant-ph \\
Quantitative Biology (Quant. Bio.) & q-bio \\
Quantitative Finance (Quant. Fin.) & q-fin \\
Statistics & stat \\
\bottomrule
\end{tabular}%
}
\caption{Mapping from domains to arXiv category labels.}
\label{table:domain_mapping}
\end{table}

\section{Further Quantitative Analysis}
\label{appendix:quantitative_analysis}
In this section, we provide a more granular, domain-by-domain breakdown of our quantitative results to better understand the performance characteristics of each model across different scientific disciplines. 

Table \ref{table:domain_results} presents the average similarity scores across the entirety of \ourbenchmark{}. The results demonstrate that \ourmethod{} achieves consistent, robust improvements over the other methods. Notably, this performance gap holds steady across highly diverse fields, ranging from largely theoretical domains like Mathematics and Theory to applied disciplines such as Vision, Robotics, and HCI.

To further validate the generalization capabilities of our approach and ensure the model is not merely memorizing training distributions, we separately evaluated performance on a highly constrained subset of \ourbenchmark{}: \textbf{Test-unseen-parents}. This subset exclusively contains test examples where the parent papers were never encountered during the training phase. 

The results for this holdout set are detailed in Table \ref{table:domain_results_unseen}. Encouragingly, \ourmethod{} maintains its substantial performance gap over the baselines even on unseen literature. This indicates that our model has successfully learned generalized mechanisms for cross-paper insight synthesis that translate effectively to novel scientific texts.
\begin{table}[H]
\centering
\resizebox{0.95\textwidth}{!}{%
\begin{tabular}{lcccccc}
\toprule
\textbf{Domain} & \textbf{Base} & \textbf{SFT} & \textbf{SFT-think} & \textbf{\gemtwopro{}} & \textbf{\gemthreepro{}} & \textbf{\ourmethod{}} \\
\midrule
Economics & $4.66_{{\pm0.21}}$ & $4.78_{{\pm0.20}}$ & $4.94_{{\pm0.21}}$ & $4.75_{{\pm0.22}}$ & $4.50_{{\pm0.20}}$ & \textbf{$5.47_{{\pm0.22}}$} \\
Language & $4.50_{{\pm0.10}}$ & $4.75_{{\pm0.10}}$ & $4.84_{{\pm0.11}}$ & $4.45_{{\pm0.11}}$ & $4.14_{{\pm0.10}}$ & \textbf{$5.50_{{\pm0.11}}$} \\
Vision & $4.28_{{\pm0.10}}$ & $4.59_{{\pm0.10}}$ & $4.57_{{\pm0.10}}$ & $4.11_{{\pm0.10}}$ & $4.14_{{\pm0.10}}$ & \textbf{$5.52_{{\pm0.12}}$} \\
Theory & $4.54_{{\pm0.10}}$ & $4.72_{{\pm0.10}}$ & $4.80_{{\pm0.11}}$ & $4.67_{{\pm0.11}}$ & $4.58_{{\pm0.11}}$ & \textbf{$5.63_{{\pm0.11}}$} \\
Mathematics & $4.64_{{\pm0.11}}$ & $4.73_{{\pm0.10}}$ & $4.81_{{\pm0.10}}$ & $4.76_{{\pm0.11}}$ & $4.58_{{\pm0.10}}$ & \textbf{$5.66_{{\pm0.11}}$} \\
Quant. Fin. & $4.82_{{\pm0.25}}$ & $5.15_{{\pm0.23}}$ & $5.31_{{\pm0.25}}$ & $5.04_{{\pm0.26}}$ & $4.45_{{\pm0.24}}$ & \textbf{$5.75_{{\pm0.26}}$} \\
ML/AI & $4.77_{{\pm0.11}}$ & $4.97_{{\pm0.11}}$ & $4.87_{{\pm0.11}}$ & $4.64_{{\pm0.11}}$ & $4.56_{{\pm0.11}}$ & \textbf{$5.76_{{\pm0.12}}$} \\
Systems & $4.62_{{\pm0.11}}$ & $5.07_{{\pm0.11}}$ & $5.13_{{\pm0.11}}$ & $4.57_{{\pm0.11}}$ & $4.41_{{\pm0.11}}$ & \textbf{$6.04_{{\pm0.12}}$} \\
Society & $4.87_{{\pm0.14}}$ & $5.31_{{\pm0.14}}$ & $5.33_{{\pm0.15}}$ & $4.40_{{\pm0.13}}$ & $3.82_{{\pm0.11}}$ & \textbf{$6.05_{{\pm0.15}}$} \\
Physics & $4.82_{{\pm0.11}}$ & $5.27_{{\pm0.11}}$ & $5.30_{{\pm0.11}}$ & $4.95_{{\pm0.11}}$ & $4.65_{{\pm0.11}}$ & \textbf{$6.14_{{\pm0.12}}$} \\
Robotics & $4.74_{{\pm0.11}}$ & $5.06_{{\pm0.11}}$ & $5.11_{{\pm0.11}}$ & $4.55_{{\pm0.11}}$ & $4.43_{{\pm0.11}}$ & \textbf{$6.21_{{\pm0.11}}$} \\
EE \& Sys. & $4.83_{{\pm0.11}}$ & $5.02_{{\pm0.11}}$ & $5.08_{{\pm0.11}}$ & $4.74_{{\pm0.11}}$ & $4.55_{{\pm0.11}}$ & \textbf{$6.22_{{\pm0.12}}$} \\
HCI & $4.95_{{\pm0.12}}$ & $5.12_{{\pm0.11}}$ & $5.02_{{\pm0.11}}$ & $4.52_{{\pm0.11}}$ & $4.01_{{\pm0.10}}$ & \textbf{$6.23_{{\pm0.12}}$} \\
Statistics & $5.45_{{\pm0.16}}$ & $5.38_{{\pm0.15}}$ & $5.51_{{\pm0.15}}$ & $5.50_{{\pm0.16}}$ & $5.68_{{\pm0.17}}$ & \textbf{$6.46_{{\pm0.16}}$} \\
Quant. Bio. & $5.17_{{\pm0.17}}$ & $5.43_{{\pm0.17}}$ & $5.69_{{\pm0.17}}$ & $4.94_{{\pm0.17}}$ & $4.41_{{\pm0.15}}$ & \textbf{$6.65_{{\pm0.17}}$} \\
CS-Other & $5.16_{{\pm0.16}}$ & $5.58_{{\pm0.16}}$ & $5.42_{{\pm0.16}}$ & $4.57_{{\pm0.15}}$ & $4.19_{{\pm0.14}}$ & \textbf{$6.70_{{\pm0.16}}$} \\
\midrule
\textbf{Overall} & $4.75_{{\pm0.03}}$ & $5.01_{{\pm0.03}}$ & $5.05_{{\pm0.03}}$ & $4.65_{{\pm0.03}}$ & $4.43_{{\pm0.03}}$ & \textbf{$5.97_{{\pm0.03}}$} \\
\bottomrule
\end{tabular}%
}
\caption{Average similarity scores (mean $\pm$ standard error) per domain on \ourbenchmark{}. The judge LM is \gemthreepro{}.}
\label{table:domain_results}
\end{table}

\begin{table}[H]
\centering
\resizebox{0.95\textwidth}{!}{%
\begin{tabular}{lcccccc}
\toprule
\textbf{Domain} & \textbf{Base} & \textbf{SFT} & \textbf{SFT-think} & \textbf{\gemtwopro{}} & \textbf{\gemthreepro{}} & \textbf{\ourmethod{}} \\
\midrule
Economics & $4.63_{{\pm0.22}}$ & $4.73_{{\pm0.21}}$ & $4.93_{{\pm0.21}}$ & $4.76_{{\pm0.23}}$ & $4.55_{{\pm0.21}}$ & \textbf{$5.50_{{\pm0.23}}$} \\
Language & $4.59_{{\pm0.19}}$ & $4.79_{{\pm0.19}}$ & $4.96_{{\pm0.20}}$ & $4.52_{{\pm0.19}}$ & $4.28_{{\pm0.18}}$ & \textbf{$5.62_{{\pm0.20}}$} \\
Vision & $4.45_{{\pm0.14}}$ & $4.57_{{\pm0.15}}$ & $4.64_{{\pm0.14}}$ & $4.14_{{\pm0.14}}$ & $4.33_{{\pm0.15}}$ & \textbf{$5.71_{{\pm0.17}}$} \\
Theory & $4.63_{{\pm0.12}}$ & $4.80_{{\pm0.12}}$ & $4.76_{{\pm0.12}}$ & $4.84_{{\pm0.13}}$ & $4.75_{{\pm0.13}}$ & \textbf{$5.71_{{\pm0.13}}$} \\
Mathematics & $4.63_{{\pm0.11}}$ & $4.72_{{\pm0.10}}$ & $4.80_{{\pm0.10}}$ & $4.74_{{\pm0.11}}$ & $4.57_{{\pm0.10}}$ & \textbf{$5.63_{{\pm0.11}}$} \\
Quant. Fin. & $4.76_{{\pm0.27}}$ & $5.18_{{\pm0.25}}$ & $5.30_{{\pm0.26}}$ & $4.89_{{\pm0.27}}$ & $4.48_{{\pm0.25}}$ & \textbf{$5.77_{{\pm0.27}}$} \\
ML/AI & $4.88_{{\pm0.15}}$ & $5.18_{{\pm0.16}}$ & $4.95_{{\pm0.15}}$ & $4.87_{{\pm0.16}}$ & $4.99_{{\pm0.16}}$ & \textbf{$6.06_{{\pm0.16}}$} \\
Systems & $4.74_{{\pm0.15}}$ & $5.04_{{\pm0.15}}$ & $5.07_{{\pm0.15}}$ & $4.74_{{\pm0.16}}$ & $4.54_{{\pm0.15}}$ & \textbf{$6.15_{{\pm0.16}}$} \\
Society & $4.96_{{\pm0.19}}$ & $5.45_{{\pm0.19}}$ & $5.26_{{\pm0.19}}$ & $4.47_{{\pm0.17}}$ & $3.77_{{\pm0.14}}$ & \textbf{$6.13_{{\pm0.20}}$} \\
Physics & $4.82_{{\pm0.11}}$ & $5.28_{{\pm0.11}}$ & $5.29_{{\pm0.11}}$ & $4.99_{{\pm0.12}}$ & $4.67_{{\pm0.11}}$ & \textbf{$6.16_{{\pm0.12}}$} \\
Robotics & $4.89_{{\pm0.13}}$ & $5.14_{{\pm0.12}}$ & $5.19_{{\pm0.13}}$ & $4.58_{{\pm0.13}}$ & $4.48_{{\pm0.13}}$ & \textbf{$6.34_{{\pm0.13}}$} \\
EE \& Sys. & $4.90_{{\pm0.13}}$ & $5.21_{{\pm0.13}}$ & $5.20_{{\pm0.12}}$ & $4.80_{{\pm0.13}}$ & $4.72_{{\pm0.13}}$ & \textbf{$6.35_{{\pm0.14}}$} \\
HCI & $5.19_{{\pm0.15}}$ & $5.23_{{\pm0.14}}$ & $5.15_{{\pm0.14}}$ & $4.74_{{\pm0.14}}$ & $4.17_{{\pm0.12}}$ & \textbf{$6.41_{{\pm0.15}}$} \\
Statistics & $5.47_{{\pm0.17}}$ & $5.43_{{\pm0.15}}$ & $5.53_{{\pm0.16}}$ & $5.49_{{\pm0.17}}$ & $5.67_{{\pm0.18}}$ & \textbf{$6.51_{{\pm0.16}}$} \\
Quant. Bio. & $5.27_{{\pm0.18}}$ & $5.56_{{\pm0.18}}$ & $5.77_{{\pm0.18}}$ & $5.05_{{\pm0.18}}$ & $4.35_{{\pm0.16}}$ & \textbf{$6.77_{{\pm0.18}}$} \\
CS-Other & $5.38_{{\pm0.19}}$ & $5.65_{{\pm0.18}}$ & $5.50_{{\pm0.18}}$ & $4.66_{{\pm0.17}}$ & $4.20_{{\pm0.16}}$ & \textbf{$6.85_{{\pm0.18}}$} \\
\midrule
\textbf{Overall} & $4.87_{{\pm0.04}}$ & $5.10_{{\pm0.04}}$ & $5.11_{{\pm0.04}}$ & $4.78_{{\pm0.04}}$ & $4.57_{{\pm0.04}}$ & \textbf{$6.11_{{\pm0.04}}$} \\
\bottomrule
\end{tabular}%
}
\caption{Average similarity scores (mean $\pm$ standard error) per domain on  \textbf{Test-unseen-parents}, which indicates the subset of \ourbenchmark{} test set with only unseen parent papers. The judge LM is \gemthreepro{}.}
\label{table:domain_results_unseen}
\end{table}

\newpage
\section{Prompts}

This section details the exact prompt templates utilized throughout our pipeline for both generation and evaluation tasks. 

\subsection{Insight Anticipation Prompt}
\label{appendix:insight_gen_prompt}

This is the prompt used for the insight anticipation task. The model is provided with the textual summaries of two parent papers, which are highlighted in red in the following figure. The prompt instructs the model to analyze these summaries, synthesize their core methodologies or findings, and generate a scientific insight or research direction that conceptually bridges and builds upon both foundational works.

\begin{figure}[H]
    \centering
    \includegraphics[width=\linewidth]{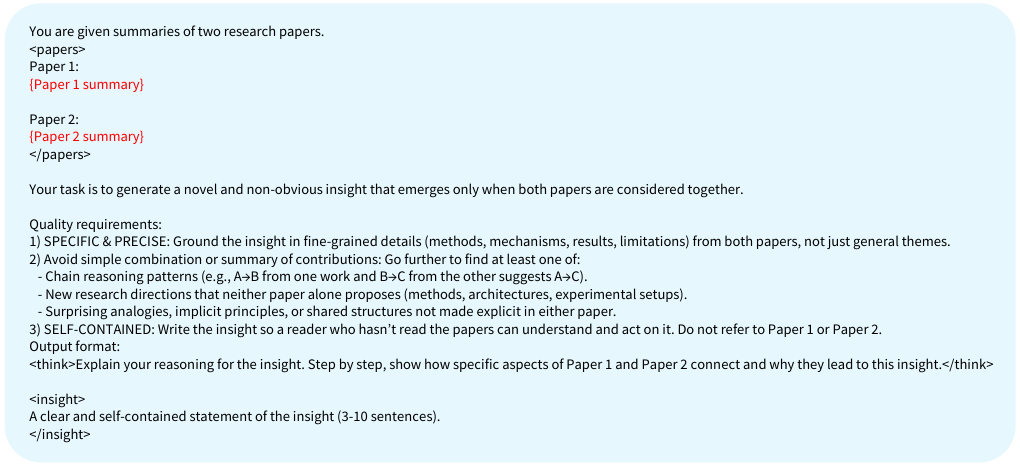}
    \caption{Prompt for insight anticipation. The summaries of two parent papers are in \textcolor{red}{red}.}
    \label{fig:insight_generation_prompt}
\end{figure}

\newpage
\subsection{Parents Identification Prompt}
\label{appendix:ancestor_id_prompt}

We present the prompt designed for the parents identification task. The primary objective of this prompt is to instruct the model to determine the most influential parent papers for a given downstream target paper. By providing the model with the downstream paper PDF, the prompt asks the model to trace the scientific lineage backward and identify which prior works from a candidate pool served as the conceptual ancestors of an insight drawn from the downstream paper.

\begin{figure}[H]
    \centering
    \includegraphics[width=\linewidth]{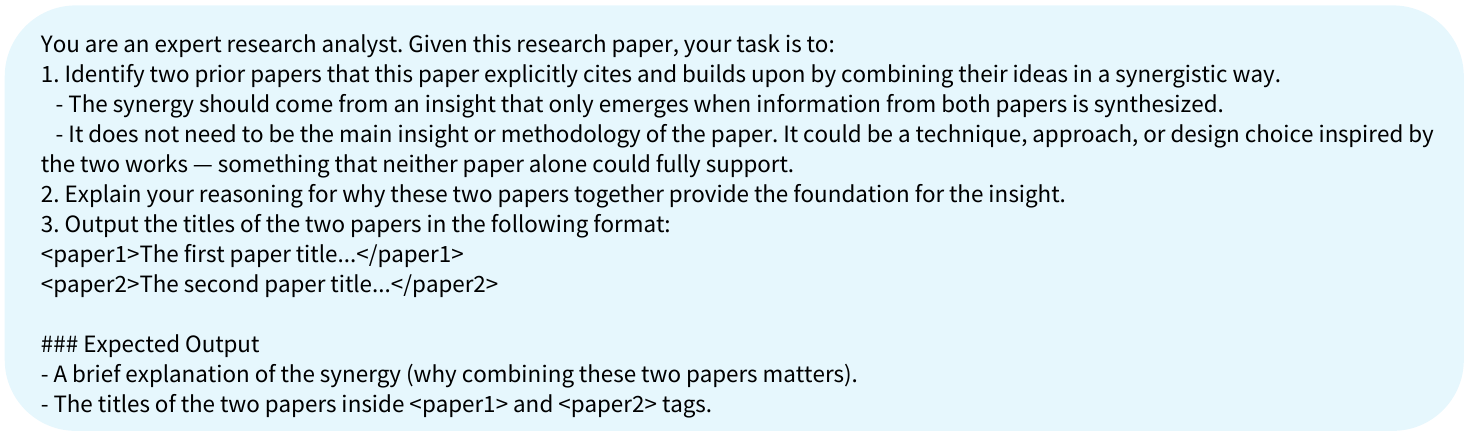}
    \vspace{-3mm}
    \caption{Prompt for identifying two parent papers whose ideas are synergistically combined to produce the given paper's key insight, which is extracted as the ground-truth target $y^*$.}
    \label{fig:upstream_citation_prompt}
\end{figure}

\newpage
\subsection{Similarity Judge Prompt}
\label{appendix:similarity_judge_prompt}

Below, we present the prompt used for automated evaluation and reward scoring via an LM-as-a-judge framework. To quantitatively assess the quality and relevance of our generated insights, this prompt provides the evaluator model with both the ground-truth insight and the corresponding model-generated insight. The model is then instructed to critically compare the two texts and output a similarity assessment, evaluating them on semantic alignment.

\begin{figure}[H]
    \centering
    \includegraphics[width=0.9\linewidth]{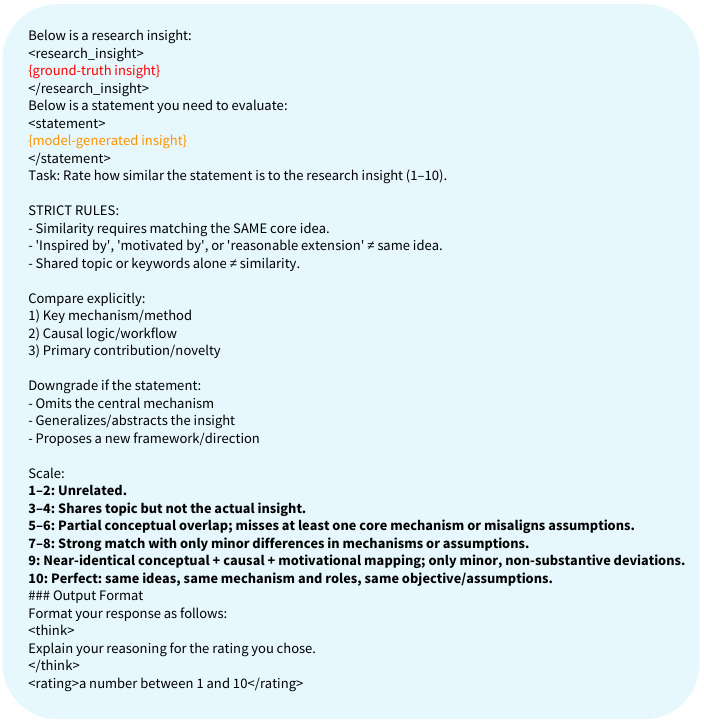}
    \caption{Prompt for assessing the similarity between the ground-truth insight $y^*$ (in \textcolor{red}{red}) and model-generated insight $\hat{y}$ (in \textcolor{orange}{orange}). The scale is \textbf{bolded}.}
    \label{fig:similarity_judge_prompt}
\end{figure}

\subsection{Rewriting Insight Prompt}
\label{appendix:rewriting_insight_prompt}

\begin{figure}[H]
    \centering
    \includegraphics[width=\linewidth]{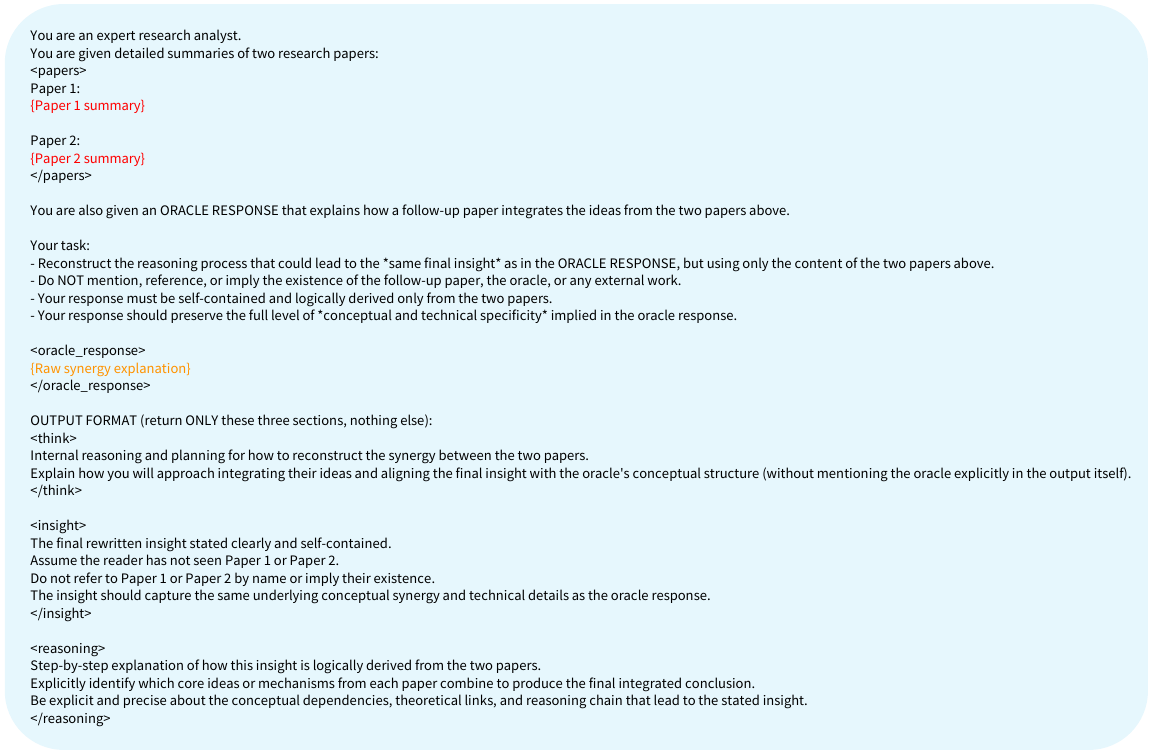}
    \caption{Prompt for rewriting the insight as a standalone statement. The summaries of two parent papers are in \textcolor{red}{red}. The raw synergy explanation is in \textcolor{orange}{orange}.}
    \label{fig:rewrite_insight_prompt}
\end{figure}

To ensure the generated insights could be evaluated consistently, we employed a rewriting step to standardize their format. Figure \ref{fig:rewrite_insight_prompt} displays the prompt used to instruct the model to rewrite the raw insight into a clear, standalone statement without losing its original semantic meaning.

\newpage
\section{Hyperparameters for \ourmethod{} and Baselines}
\label{app:hyperparameters}

We detail the optimization and training configurations used for training \ourmethod{} and the SFT baselines below. The hyperparameter choices were guided by standard practices for aligning large language models, with slight adjustments made to accommodate our specific sequence lengths and batch sizes.

\paragraph{RL Hyperparameters.} 
Table \ref{tab:rl_hyperparameters} outlines the configuration used during the reinforcement learning phase, specifically utilizing the GRPO algorithm. We applied a conservative KL penalty to prevent the model from drifting too far from the reference policy.

\begin{table*}[h]
\centering
\small
\setlength{\tabcolsep}{10pt}
\renewcommand{\arraystretch}{1.15}
\begin{tabularx}{\linewidth}{>{\raggedright\arraybackslash}X >{\centering\arraybackslash}p{0.34\linewidth}}
\toprule
\textbf{Hyperparameter} & \textbf{Value} \\
\midrule
Algorithm                           & GRPO~\citep{shao2024deepseekmath} \\
Training steps                      & 400 \\
Train batch size                    & 64 \\
Group size                          & 8 \\
Max prompt length                   & 3000 \\
Max response length                 & 1296 \\
Learning rate                       & $1\times10^{-6}$ \\
Entropy coefficient                 & 0.002 \\
KL loss coefficient                 & 0.001 \\
KL loss type                        & \texttt{low\_var\_kl} \\
Sampling temperature (train / val)  & 0.6 / 0.6 \\
Max batched tokens                  & 32768 \\
Uneven Clipping (low / high)        & 0.2 / 0.5 \\
\bottomrule
\end{tabularx}
\caption{Key reinforcement learning hyperparameters used in our experiments.}
\label{tab:rl_hyperparameters}
\end{table*}

\paragraph{Supervised Learning Hyperparameters.}
Table \ref{tab:sft_hyperparameters} details the setup for our supervised fine-tuning (SFT) baselines. 

\begin{table*}[h]
\centering
\small
\setlength{\tabcolsep}{10pt}
\renewcommand{\arraystretch}{1.15}
\begin{tabularx}{\linewidth}{>{\raggedright\arraybackslash}X >{\centering\arraybackslash}p{0.34\linewidth}}
\toprule
\textbf{Hyperparameter} & \textbf{Value} \\
\midrule
Algorithm                           & SFT \\
Epochs                              & 10 \\
Train batch size                    & 64 \\
Max length (input + output)         & 8192 \\
Learning rate                       & $1\times10^{-6}$ \\
Gradient checkpointing              & True \\
\bottomrule
\end{tabularx}
\caption{Key supervised fine-tuning hyperparameters used in our experiments. }
\label{tab:sft_hyperparameters}
\end{table*}

\section{Ablations with Diverse LM Judges}
\label{appendix:ablations_with_diverse_lm_judges}

\begin{figure}[H]
    \centering
    \includegraphics[width=\linewidth]{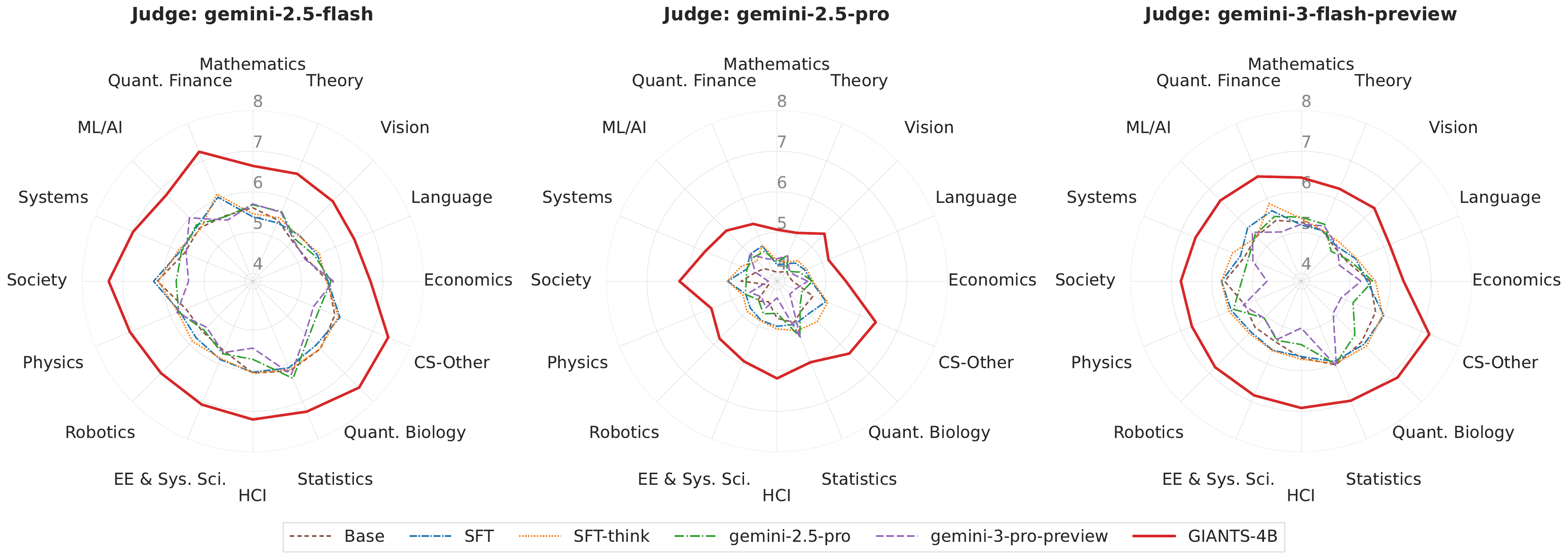}
    \vspace{-3mm}
    \caption{\textbf{Similarity scores on \emph{Test-unseen-parents}}. This is the subset of \ourbenchmark{} test set with only unseen parent papers. Across all three LM judges, \ourmethod{} is ranked as the top-performing model.}
    \label{fig:all_domain_radar_3judges_unseen_parents}
\end{figure}

\newpage
\section{Human Evaluation Details}
\label{app:human_eval}

We conduct two preliminary human studies, each involving two PhD students in Computer Science as annotators. 

\subsection{Human Evaluation of Insight Similarity}
\label{app:human_eval_similarity}

\begin{figure}[H]
    \centering
    \includegraphics[width=\linewidth]{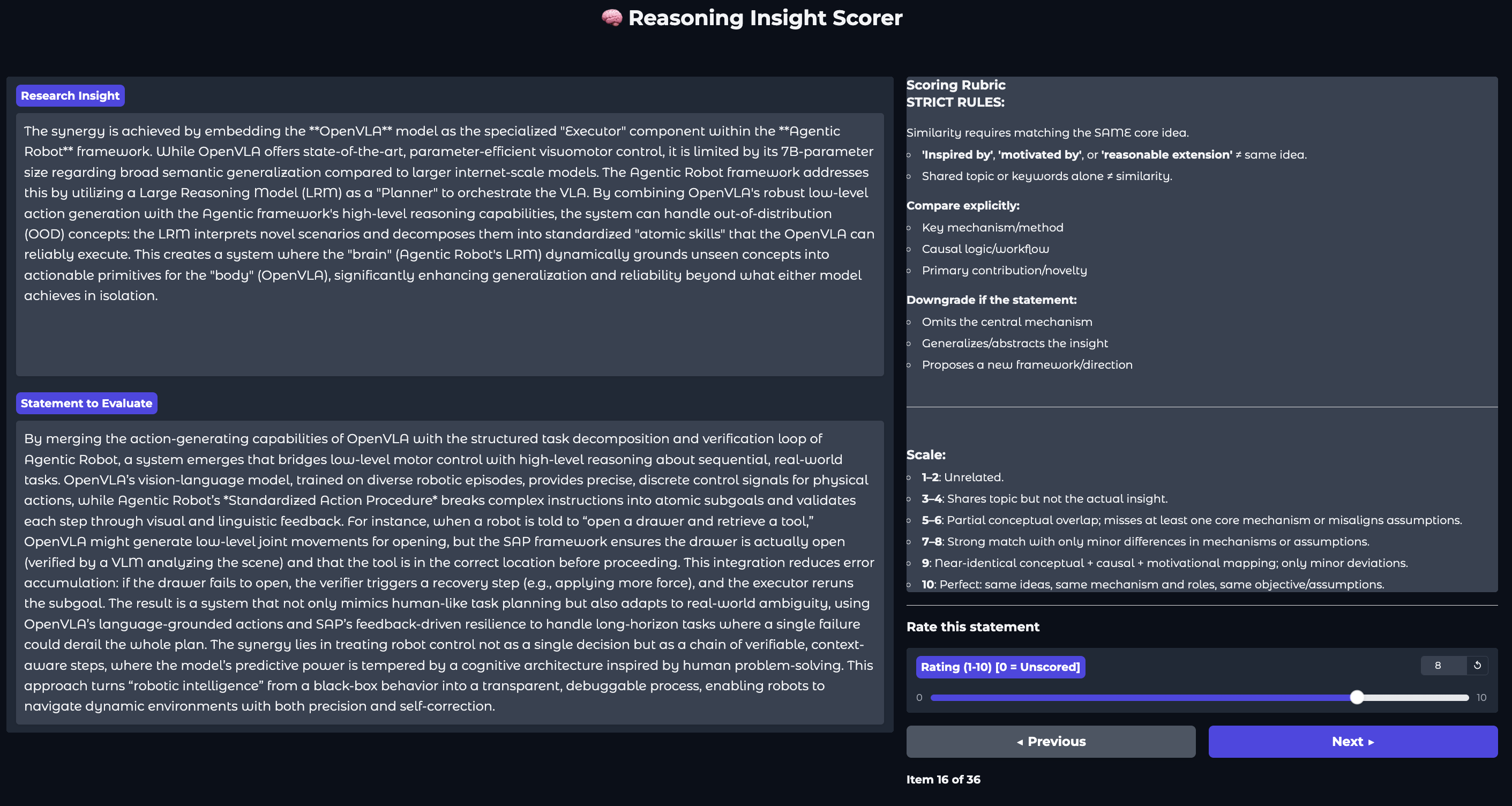}
    \caption{\textbf{Human eval labeling interface for assessing similarity.} This is the Gradio interface that was used by human annotators to measure the consistency of LM similarity scores with human ratings. The rubric provided to the humans exactly matched that provided to the LM judge for consistency (with markdown formatting for human interpretability).}
    \label{fig:human_eval}
\end{figure}

To validate the reliability of our automated metrics, we conducted a preliminary human evaluation study focusing on insight similarity. Two annotators were tasked with scoring the semantic alignment between model-generated insights and ground-truth insights. To facilitate this process and ensure an unbiased environment, we developed a custom web interface, shown in Figure \ref{fig:human_eval}. Importantly, the human annotators were provided with the exact same scoring rubric and guidelines that were given to the LM judge, allowing us to accurately measure the consistency between human judgment and our automated evaluation pipeline, with the exact context provided to the LM judge.

Both annotators rated the same 30 pairs of insights generated by the base model and \ourmethod{} against their corresponding ground-truth insights. For each insight, we compute the average similarity score across the two annotators. To compute the win rate under human judge, we compare the averaged scores of the base model and \ourmethod{} for each pair and ignore ties. 

\subsection{Human Evaluation of Insight Feasibility}
\label{app:human_eval_feasibility}
\begin{figure}[H]
    \centering
    \includegraphics[width=\linewidth]{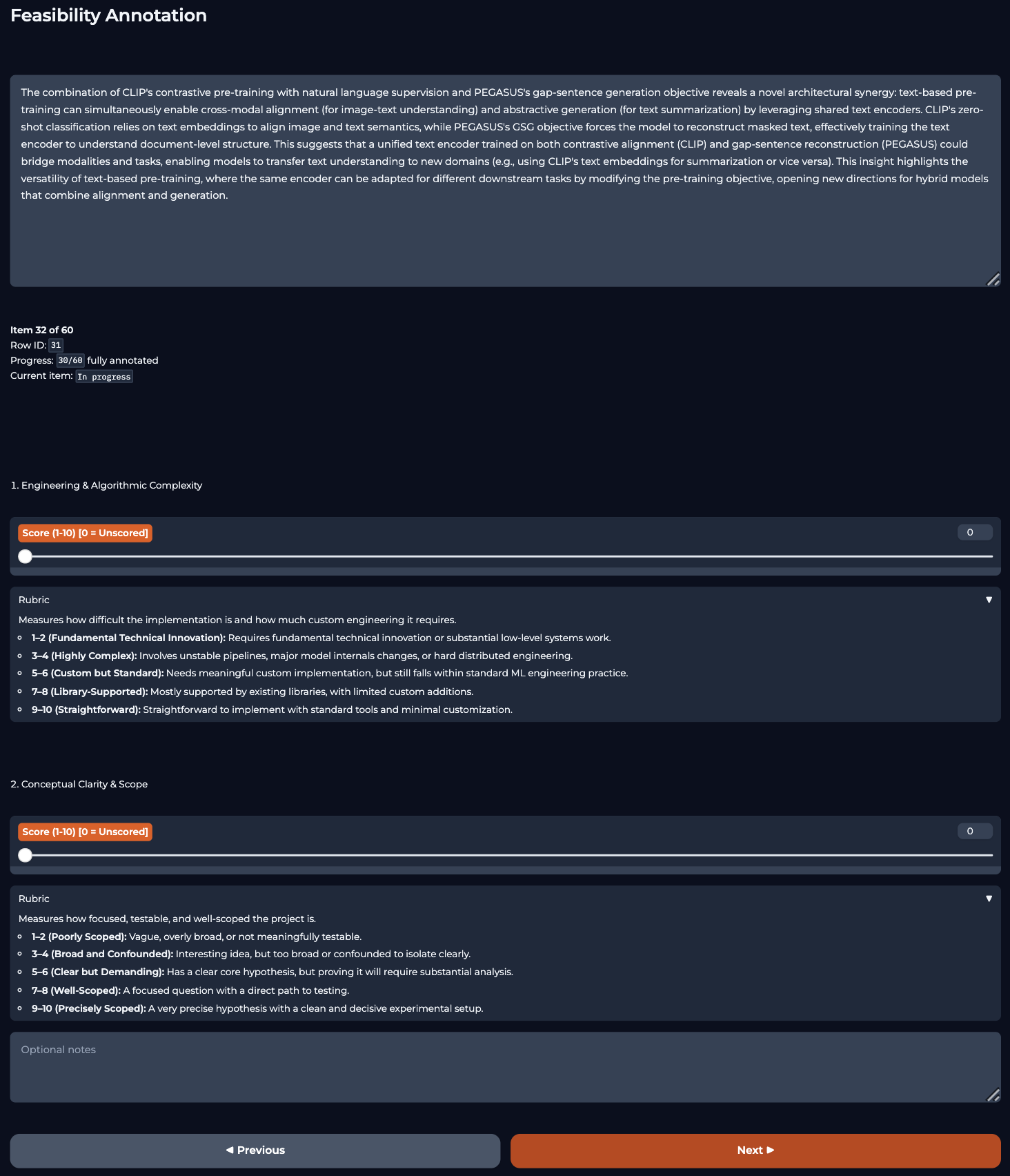}
    \caption{\textbf{Human eval labeling app for assessing feasibility.} This is the Gradio interface that was used by human annotators to measure the feasibility of a research insight along two dimensions: engineering/algorithmic complexity and conceptual clarity.}
    \label{fig:feasibility_human_eval}
\end{figure}

In the study on insight feasibility, both annotators rated the same 15 pairs of insights generated by the base model and GIANTS-4B along two dimensions: algorithmic complexity and conceptual clarity. For each insight and each dimension, we report the average rating across the two annotators. Figure~\ref{fig:feasibility_human_eval} shows the rubric for the two dimensions.

\end{document}